%% file: main.tex
\documentclass[10pt]{article} 
\usepackage[preprint]{tmlr}

\input{math_commands.tex}

\usepackage[utf8]{inputenc} 
\usepackage[T1]{fontenc}    
\usepackage[hidelinks]{hyperref}       
\usepackage{url}            
\usepackage{booktabs}       
\usepackage{amsfonts}  
\usepackage{nicefrac}       
\usepackage{microtype}      
\usepackage{xcolor}
\usepackage{graphicx}
\usepackage{caption}
\usepackage{subcaption}
\usepackage{algorithm}
\usepackage{algpseudocode}
\usepackage{wrapfig}
\usepackage{lipsum}
\usepackage{amssymb}
\usepackage{amsmath}
\usepackage{natbib}
\usepackage{colortbl}
\usepackage{cellspace}

\newcommand{\metapg}{MetaPG }
\newcommand{\metapgnospace}{MetaPG}
\newcommand{\Generalizability} {Generalizability }
\newcommand{\Generalizabilitynospace}{Generalizability} 
\newcommand{\generalizability}{generalizability }
\newcommand{\generalizabilitynospace}{generalizability}

\title{Evolving Pareto-Optimal Actor-Critic Algorithms \\ for \Generalizability and Stability}

\author{\name Juan Jose Garau-Luis$^{1}$, Yingjie Miao$^{2}$, John D. Co-Reyes$^{2}$, Aaron Parisi$^{2}$\\
        Jie Tan$^{2}$, Esteban Real$^{2}$, Aleksandra Faust$^{2}$ \\
        \addr $^{1}$MIT, $^{2}$Google Brain \\ \\
         garau@mit.edu \\
         \{yingjiemiao,parisi,jietan,ereal,faust\}@google.com}

\def\month{MM}  
\def\year{YYYY} 
\def\openreview{\url{https://openreview.net/forum?id=XXXX}} 

\begin{document}

\maketitle

\begin{abstract}
\Generalizability and stability are two key objectives for operating reinforcement learning (RL) agents in the real world. Designing RL algorithms that optimize these objectives can be a costly and painstaking process. This paper presents \metapgnospace, an evolutionary method for automated design of actor-critic loss functions. \metapg explicitly optimizes for generalizability and performance, and implicitly optimizes the stability of both metrics. We initialize our loss function population with Soft Actor-Critic (SAC) and perform multi-objective optimization using fitness metrics encoding single-task performance, zero-shot \generalizability to unseen environment configurations, and stability across independent runs with different random seeds. On a set of continuous control tasks from the Real-World RL Benchmark Suite, we find that our method, using a single environment during evolution, evolves algorithms that improve upon SAC's performance and \generalizability by 4\% and 20\%, respectively, and reduce instability up to 67\%. Then, we scale up to more complex environments from the Brax physics simulator and replicate \generalizability tests encountered in practical settings, such as different friction coefficients. \metapg evolves algorithms that can obtain 10\% better \generalizability without loss of performance within the same meta-training environment and obtain similar results to SAC when doing cross-domain evaluations in other Brax environments. The evolution results are interpretable; by analyzing the structure of the best algorithms we identify elements that help optimizing certain objectives, such as regularization terms for the critic loss.
\end{abstract}

\section{Introduction}
\label{sec:introduction}




\begin{wrapfigure}{R}{0.3\textwidth}
\begin{center}
\includegraphics[width=0.3\textwidth]{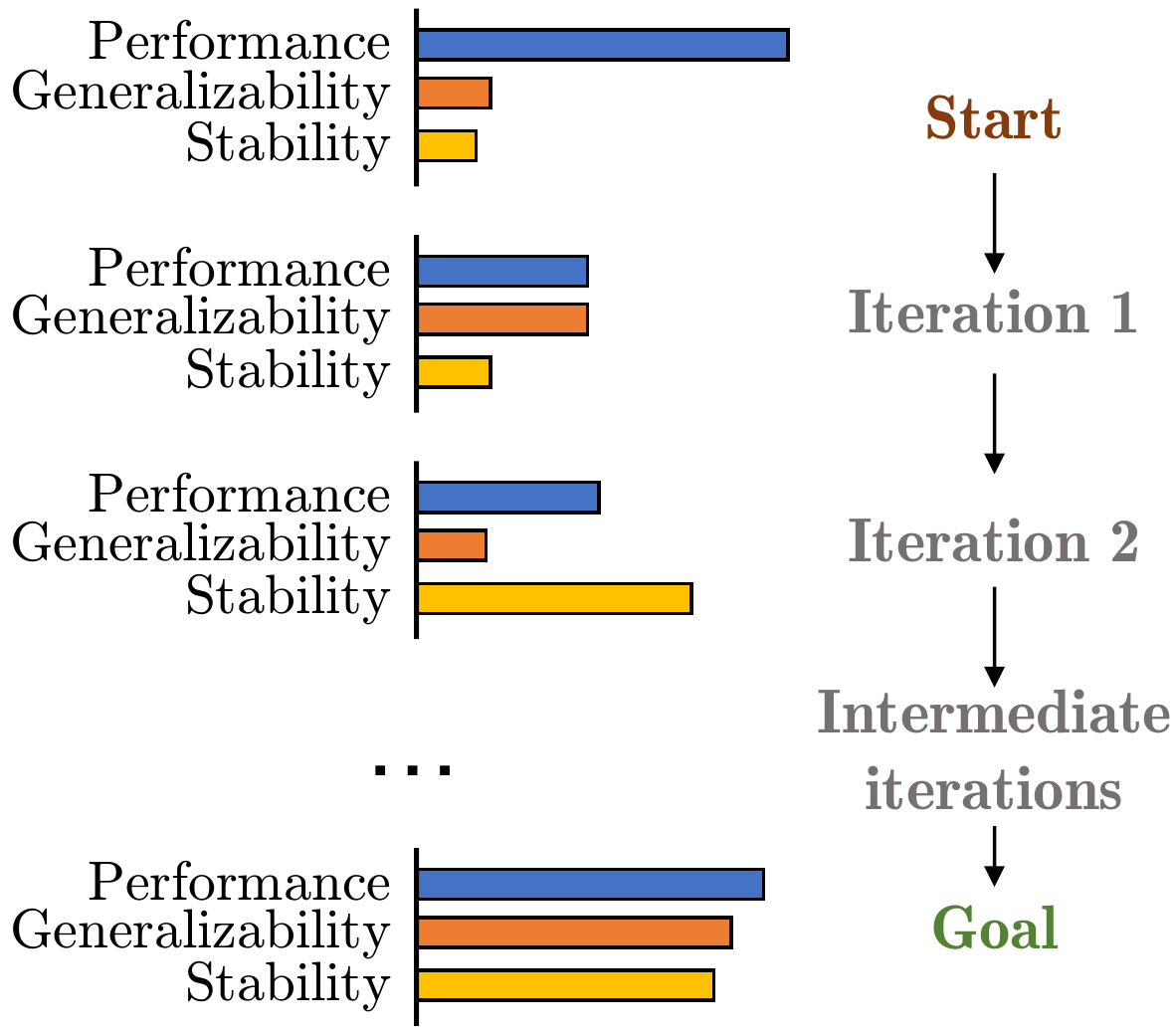}
\end{center}
\caption{\small In many practical contexts, designing the RL agent that achieves the right performance, \generalizabilitynospace, and stability is an empirical and costly iteration-based process.}
\label{fig:tradeoff_iterations}
\end{wrapfigure}
Two key bottlenecks for the deployment of reinforcement learning (RL) in the real world are failing to generalize beyond the training distribution and unstable training. Both are common in practical settings and constitute two important aspects of RL robustness. On one hand, many real-world environments present themselves in multiple configurations (e.g., different sizes, structure, context, properties) and practitioners expect zero-shot generalization when facing new configurations in robot manipulation \citep{ibarz2021train}, navigation \citep{autorl-nav}, energy systems \citep{PERERA2021110618}, and fluid dynamics \citep{garnier2021review}. On the other hand, real-world elements such as stochastic dynamics should not result in unstable learning behaviors that lead to undesired performance drops. Even for state-of-the-art RL algorithms, zero-shot generalization and instability are considerable challenges \citep{henderson2018deep,dulac2021challenges}.

Improving \generalizability has been addressed by learning or encoding inductive biases in RL algorithms \citep{raileanu2021decoupling,vlastelica2021neuro}. Gains mostly come by manually modifying existing algorithms \citep{cobbe2019quantifying,igl2019generalization,cobbe2021phasic}. As RL design for real-world environments tends to be empirical \citep{andrychowicz2020matters,linke2020adapting}, finding the suitable algorithmic changes that benefit \generalizability might take many iterations, especially if performance loss must be avoided (see Figure \ref{fig:tradeoff_iterations}). As environments become more complex and inductive biases become environment-specific, the cost of human-driven design might be too expensive when optimizing for generalizability \citep{zhao2019investigating}, let alone when optimizing for both performance and \generalizability \citep{hessel2019inductive}, which are just two objectives we consider but there are many more.
In addition, stable algorithms that show consistent performance and \generalizability across independent runs leads to higher algorithm reusability within the same environment and across environments. We argue that optimizing \generalizability and stability in addition to performance in RL builds the case for automating algorithm design and speeding up the process of RL algorithm discovery. Automated Machine Learning or AutoML \citep{hutter2019automated} has proven to be a successful tool for supervised learning problems \citep{vinyals2016matching,zoph2018learning,real2019regularized,finn2017model}, and it has been recently applied in the context of RL for automating loss function search \citep{Oh2020,Xu2020,co2021evolving,Bechtle2021,he2022reinforcement,lu2022discovered}.

\begin{figure}[t]
\begin{center}
\includegraphics[width=0.95\textwidth]{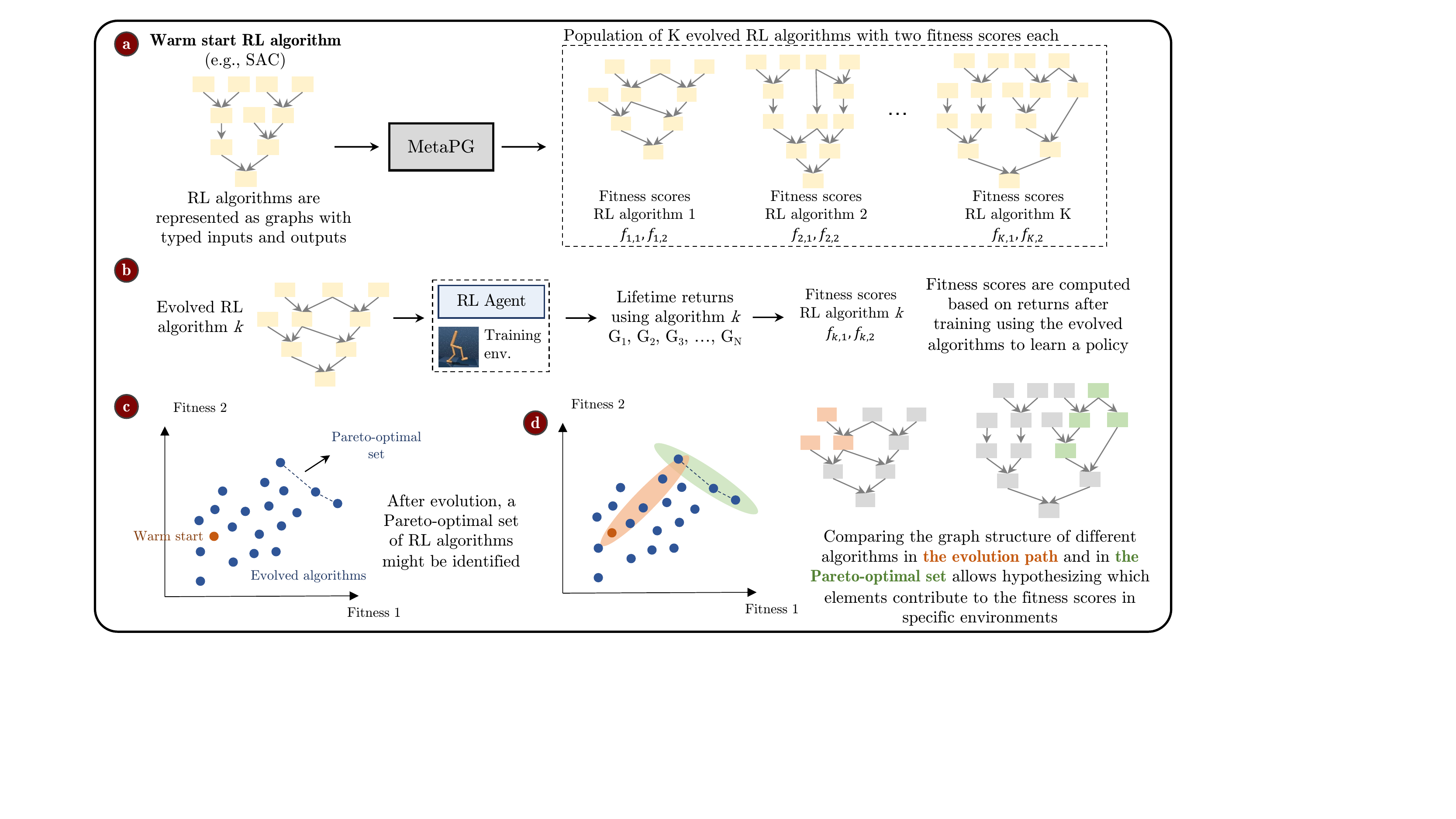}
\end{center}
\caption{\small \metapg overview, example with two fitness scores encoding two RL objectives. \textbf{(a)} The method starts by taking a warm-start RL algorithm with its loss function represented in the form of a directed acyclic graph. \metapg consists of a meta-evolution process that, after initializing algorithms to the warm-start, discovers a population of new algorithms. \textbf{(b)} Each evolved graph is evaluated by training an agent following the algorithm encoded by it, and then computing two fitness scores based on the training outcome. \textbf{(c)} After evolution, all RL algorithms can be represented in the fitness space and a Pareto-optimal set of algorithms can be identified. \textbf{(d)} Identifying which graph substructures change across the algorithms in the Pareto set reveals which operations are useful for specific RL objectives. \metapg can be scaled to more than two RL objectives.}
\label{fig:metapg_diagram}
\end{figure}

This paper proposes \metapg (see Figure \ref{fig:metapg_diagram}), a method that evolves a population of actor-critic RL algorithms \citep{sutton2018reinforcement}, identified by their loss functions, with the goal of increasing single-task performance, zero-shot \generalizabilitynospace, and stability across independent runs. Loss functions are represented symbolically as directed acyclic computation graphs and two independent fitness scores are used to encode performance and \generalizabilitynospace, respectively. A measure for stability is accounted for in both objectives and the multi-objective ranking algorithm NSGA-II \citep{deb2002fast} is used to identify the best algorithms. Compared to manual design, this strategy allows us to explore the algorithm space more efficiently by automating search operations. MetaPG finds algorithm improvement directions that jointly optimize both objectives until it obtains a Pareto Front of loss functions that maximizes fitness with respect to each objective, approximating the underlying tradeoff between them. On one end of the Pareto Front we obtain algorithms that perform well in the training task (beneficial when faster learning is required and overfitting is not a concern), on the other end we find algorithms that generalize better to unseen configurations, and in between there are algorithms that interpolate between both behaviors.

To evaluate \metapgnospace, we run experiments using, first the zero-shot \generalizability benchmark from the Real-World RL environment suite \citep{dulac2021challenges}; and second, the Ant and Humanoid environments from Brax physics simulator \cite{brax2021github}, where we simulate perturbations like mass changes, different friction coefficients, and joint torques. We warm-start the evolution with a graph-based representation of Soft Actor-Critic (SAC) \citep{haarnoja2018soft}, and demonstrate that our method is able to evolve a Pareto Front of multiple actor-critic algorithms that outperform SAC and increase its performance and \generalizability up to 4\% and 20\%, respectively, and decrease instability up to 67\%. 
In the Brax environments, the evolved algorithms outperform SAC by 15\% and 10\% in performance and \generalizabilitynospace, respectively. Instability is reduced 23\% in that case. Furthermore, we observe that algorithms evolved in Ant show minimal \generalizability loss when transferred to Humanoid and vice versa. Finally, by inspecting the graphs of the evolved algorithms, we interpret which substructure drive the gains. For instance, we find that \metapg evolves loss functions that remove the entropy term in SAC to increase performance.

In summary, this paper makes three main contributions:
\begin{enumerate}
    \item A method that combines multi-objective evolution with a search language representing actor-critic RL algorithms as graphs, which can discover new loss functions over a set of different objectives.
    \item A formulation of stability-adjusted scores that explicitly optimize for performance and \generalizability and implicitly encourage stability.
    \item A dataset\footnote{The dataset can be found at: \href{https://github.com/authors2022/dataset}{https://github.com/authors2022/dataset}} of Pareto-optimal actor-critic loss functions which outperform baselines like SAC on multiple objectives. This dataset may be further analyzed to understand how algorithmic changes affect the tradeoff between different objectives.
\end{enumerate}

\section{Related Work}
\label{sec:related_work}

\textbf{\Generalizability in RL}\quad One of the aspects of RL robustness is \generalizability \citep{kirk2022survey,xu2022trustworthy}. Increasing \generalizability has been addressed by means of environment randomization \citep{tobin2017domain,peng2018sim,akkaya2019solving}. Other authors have shown that removing or adding certain algorithmic components impacts \generalizability (e.g. using batch normalization \citep{cobbe2019quantifying}, adding elements to rewards \citep{chen2020reinforcement}, or using regularizers \citep{igl2019generalization,cobbe2019quantifying}). Others directly achieve \generalizability gains by modifying existing actor-critic RL algorithms \citep{raileanu2021decoupling,cobbe2021phasic}. \citet{vlastelica2021neuro} propose a hybrid architecture combining a neural network with a shortest path solver to find better inductive biases that improve generalizing to unseen environment configurations. We automate the search of algorithmic changes for actor-critic algorithms using \generalizability as one of the search metrics.

\textbf{Stability in RL}\quad Achieving stable behaviors during training is essential in many domains, particularly in control applications \citep{ibarz2021train,azar2021drone}. It has been shown that randomness can play a substantial role in the outcome of a training run \citep{henderson2018deep}. Stable learning has been sought by means of algorithmic innovation \citep{haarnoja2018soft,fox2015taming,bao2021deep,jiang2021emphatic}. New stable algorithms have been mainly developed after looking into stability in isolation; in this work we focus on stability as one of the three objectives to simultaneously optimize.

\textbf{Optimizing RL components}\quad Automated RL or AutoRL seeks to meta-learn RL components \citep{Parker-Holder2022}, such as RL algorithms \citep{co2021evolving,Kirsch2019,Oh2020,Bechtle2021}, their hyperparameters \citep{pmlr-v130-zhang21n, hertel2020quantity,NEURIPS2018_2715518c}, policy/neural network \citep{NEURIPS2019_e9874147,miao2021rl}, or the environment \citep{ferreira2021learning,DBLP:conf/nips/GurJMCTLF21,DBLP:conf/nips/0001JVBRCL20,pmlr-v80-florensa18a,volz2018evolving, Faust2019}. This work evolves RL loss functions and leaves other elements of the RL problem out of the scope. We focus on the RL loss function given its interaction with all elements in a RL problem: states, actions, rewards, and the policy.

\textbf{Evolutionary AutoML}\quad Neuro-evolution introduced evolutionary methods in the context of AutoML \citep{miller1989designing,stanley2002evolving}, including neural network architecture search \citep{stanley2009hypercube,jozefowicz2015empirical,real2019regularized}. In the RL context evolution searched for policy gradients \citep{houthooft2018evolved} and value iteration losses \citep{co2021evolving}. Our work is also related to the field of genetic programming, in which the goal is to discover computer code \citep{koza1994genetic,real2020automl,co2021evolving}. In this work we use a multi-objective evolutionary method to discover new RL algorithms, specifically actor-critic algorithms \citep{sutton2018reinforcement}, represented as graphs. 

\textbf{Learning RL loss functions}\quad Loss functions play a central role in RL algorithms and are traditionally designed by human experts. Recently, several lines of work propose to view RL loss functions as tunable objects that can be optimized automatically \citep{Parker-Holder2022}. One popular approach is to use neural loss functions whose parameters are optimized via meta-gradient \citep{Kirsch2019,Bechtle2021,Oh2020,Xu2020,lu2022discovered}. An alternative is to use symbolic representations of loss functions and formulate the problem as optimizing over a combinatorial space. One example is \citep{Alet2020}, which represents extrinsic rewards as a graph and optimizes it by cleverly pruning a search space. Learning value-based RL loss functions by means of evolution was first proposed by \citet{co2021evolving}, and was applied to solving discrete action problems. \citet{he2022reinforcement} propose a method to evolve auxiliary loss functions which complemented predefined standard RL loss functions. \metapg focuses on continuous control problems and searches for complete symbolic loss functions of actor-critic algorithms.

\section{Methods}
\label{sec:methods}

We represent actor-critic loss functions (policy loss and critic loss) as directed acyclic graphs and use an evolutionary algorithm to evolve a population of graphs, which are ranked based on their fitness scores. The population is warm-started with known algorithms such as SAC and undergoes mutations over time. Each graph's fitnesses are measured by training from scratch an RL agent with the corresponding loss function and encode performance and \generalizability as explicit objectives, and stability as a third objective defined implicitly. We use the multi-objective evolutionary algorithm NSGA-II \citep{deb2002fast} to jointly optimize all fitness scores until growing a Pareto-optimal set of graphs or Pareto Front. Algorithm \ref{alg:overview} summarizes the process; \texttt{Offspring} and \texttt{RankAndSelect} are NSGA-II subroutines. 

Section \ref{sec:rl-rep} provides RL algorithm graph representation details. The main logic of \metapg is contained in the evaluation routine, which computes fitness scores (Section \ref{sec:fitness}) and employs several techniques to speed up the evolution and evaluation processes (Section \ref{sec:evolution-details}). See Appendix \ref{sec:implementation_details} for further implementation details.

\begin{figure}[!h]
  \centering
  \begin{minipage}{.7\linewidth}
    \begin{algorithm}[H]
    \caption{\metapg Overview}\label{alg:overview}
    \textbf{Input:} Training environments $\mathcal{E}$ \\
    \textbf{Initialize}: Initialize population $P_0$ of loss function graphs (random initialization or bootstrap with an algorithm such as SAC).
    \begin{algorithmic}[1]
    \For{$L$ in $P_0$}
        $L.score \gets \text{Eval}(L, \mathcal{E})$
    \EndFor
    \State $Q_0 \gets$ Offspring($P_0$) \Comment{NSGA-II}
    \For{ $L$ in  $Q_0$}
        $L.score \gets \text{Eval}(L, \mathcal{E})$
    \EndFor    
    \For{$t=1$ \textbf{to} $G$}
        \State $R \gets P_{t-1} \bigcup Q_{t-1}$
        \State $P_t \gets $ RankAndSelect(R) \Comment{NSGA-II}
        \State $Q_t \gets $ Offspring($P_t$) \Comment{NSGA-II}
        \For{$L$ in $Q_t$}
            $L.score \gets \text{Eval}(L, \mathcal{E})$
        \EndFor        
    \EndFor
    \State \textbf{Output:} Pareto Front of all loss function graphs.
    \end{algorithmic}
    \end{algorithm}
  \end{minipage}
\end{figure}

\subsection{RL algorithm representation}
\label{sec:rl-rep}

\metapg encodes loss functions as graphs consisting of typed nodes sufficient to represent a wide class of actor-critic algorithms. Compared to the prior value-based RL evolutionary search method introduced by \citet{co2021evolving}, \metapgnospace's search space greatly expands on it and adds input and output types to manage the search complexity. As a representative example, Figure \ref{fig:warm_start} in Appendix \ref{sec:warm_start} presents the encoding for SAC that we use in this paper. In our experiments we limit the number of nodes per graph to 60 and 80, which can represent approximately $10^{300}$ and $10^{400}$ graphs, respectively (see Appendix \ref{sec:size_of_search_space}). Nodes in the graph encode loss function inputs, operations, and loss function outputs. The inputs include elements from transition tuples, constants such as the discount factor $\gamma$, a policy network $\pi$, and multiple critic networks $Q_i$. Operation nodes support intermediate algorithm instructions such as basic arithmetic neural network operations. Then, outputs of the graphs correspond to the policy and critic losses. The gradient descent minimization process takes these outputs and computes their gradient with respect to the respective network parameters. In Appendix \ref{sec:nodes_list} we provide a full description of the search language and nodes considered. \metapgnospace's search language supports both on-policy and off-policy algorithms; however, in this paper we focus on off-policy algorithms given their better sample efficiency.

\subsection{Fitness scores}
\label{sec:fitness}

\begin{figure}[t]
\begin{center}
\includegraphics[width=0.98\textwidth]{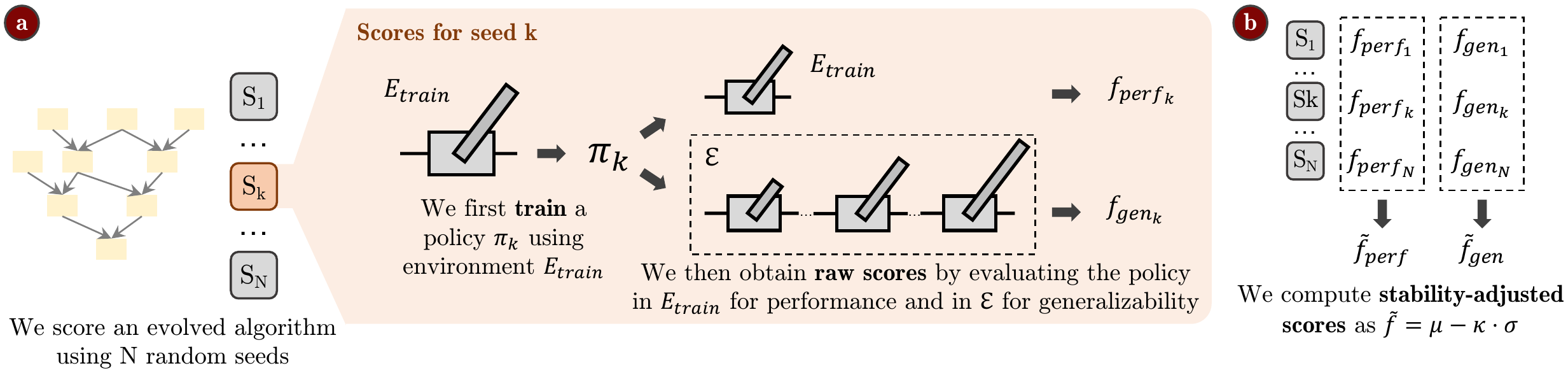}
\end{center}
\caption{\small Process to compute fitness scores. \textbf{(a)} Independently for each seed from a set of $N$ seeds, we first train a policy $\pi_k$ using environment $E_{train}$ and seed $k$. During training the policy is allowed to take stochastic actions. Then, we evaluate $\pi_k$ deterministically on that same environment $E_{train}$ to get a raw performance score $f_{perf_k}$, and on a set of environments $\mathcal{E}$ (same environment with different configurations; e.g., different pole lengths as shown) to get a raw \generalizability score $f_{gen_k}$. \textbf{(b)} We compute stability-adjusted fitness scores $\tilde{f}_{perf}$ and $\tilde{f}_{gen}$ by aggregating raw scores from each seed using Equation \ref{eq:multi-objective}.}
\label{fig:scoring_mechanism}
\end{figure}

This work focuses on optimizing single-task performance, zero-shot \generalizabilitynospace, and stability across independent runs with different random seeds. The process to compute the fitness scores is depicted in Figure \ref{fig:scoring_mechanism}. We use $N$ random seeds and a set of environments $\mathcal{E}$, which comprises multiple instances of the same environment class, including a training instance $E_{train}\in \mathcal{E}$. For example, $\mathcal{E}$ is the set of all RWRL Cartpole environments with different pole lengths (0.1 meters to 3 meters in 10-centimeter intervals), and $E_{train}$ corresponds to an instance with a specific pole length (1 meter). The first step (see Figure \ref{fig:scoring_mechanism}a) is to compute raw performance and \generalizability scores for each individual seed. Using $E_{train}$ and seed $k$ to train a policy $\pi_k$, the performance score $f_{perf_k}$ is the average evaluation return on the training environment configuration:
\begin{equation}
    \label{eq:perf}
    f_{perf_k} = \frac{1}{N_{eval}}\sum_{n=1}^{N_{eval}} G_n(\pi_k, E_{train}),
\end{equation}
where $G_n$ corresponds to the normalized return for episode $n$ given a policy and an environment instance, and $N_{eval}$ is the number of evaluation episodes. Algorithms that learn faster in the training environment and overfit to it obtain higher performance scores. The \generalizability score $f_{gen_k}$ is in turn computed as the average evaluation return of the policy trained on $E_{train}$ over the whole range of environment configurations. We emphasize that the policy is trained on a single environment configuration (for example 1.0 meter pole length) and then is evaluated in a zero-shot fashion to new unseen environment configurations:\footnote{More precisely, it should be $E \in \mathcal{E} \setminus \{E_{train}\}$. In practice, we find this makes no significant difference in the metric because the number of test configurations is normally around 30.}
\begin{equation}
    \label{eq:gen}
    f_{gen_k} = \frac{1}{|\mathcal{E}|N_{eval}}\sum_{E\in \mathcal{E}}\sum_{n=1}^{N_{eval}} G_n(\pi_k, E)
\end{equation}
We look for stable training results for both performance and \generalizabilitynospace. To that end, once we have independent raw scores for each seed ($N$ different performance and \generalizability scores) we define the \textit{stability-adjusted} scores (see Figure \ref{fig:scoring_mechanism}b) as
\begin{equation}
\label{eq:multi-objective}
\tilde{f} = \mu(\{f_n\}_{n=1}^{N}) - \kappa\cdot\sigma(\{f_n\}_{n=1}^{N})
\end{equation}
where $f$ is a score (performance or \generalizabilitynospace), $f_{n}$ denotes the score for seed $n$; $\mu$ and $\sigma$ are the mean and standard deviation across the $N$ seeds, respectively; and $\kappa$ is a penalization coefficient. The final fitness of a graph is the tuple ($\tilde{f}_{perf}$, $\tilde{f}_{gen}$).

\subsection{Evolution details}
\label{sec:evolution-details}

\textbf{Mutation}\quad The population is initialized with a warm-start RL algorithm; all individuals are copies of this algorithm's graph at the beginning. Once the population is initialized, individuals undergo mutations that change the structure of their respective graphs. Specifically, mutations consist of either replacing one or more nodes in the graph or changing the connectivity for one edge. The specific number of nodes that are affected by mutation is randomly sampled for each different individual; see Appendix \ref{sec:implementation_details} for more details.

\textbf{Operation consistency}\quad To prevent introducing corrupted child graphs into the population, \metapg checks operation consistency, i.e., for each operation, it makes sure the shapes of the input tensors are valid and compatible, and computes the shape of the output tensor. These shapes and checks are propagated along the computation graph.

\textbf{Hashing}\quad To avoid repeated evaluations, \metapg hashes \citep{real2020automl} all graphs in the population. Once the method produces a child graph and proves its consistency, it computes a hash value and, in case of a cache hit, reads the fitness scores from the cache. Since only the gradient of a loss function matters during training, we hash a graph by computing the corresponding loss function's gradient on synthetic inputs.

\textbf{Hurdle evaluations}\quad We carry out evaluations for different individuals in the population in parallel, while evaluating across seeds for one algorithm is done sequentially. To prevent spending too many resources on algorithms that are likely to yield bad policies, \metapg uses a simple hurdle environment \citep{co2021evolving} and a number of hurdle seeds. We first evaluate the algorithm on the hurdle environment for each hurdle seed, and only proceed with more complex and computationally expensive environments if the resulting policy performs above a certain threshold on the hurdle environment.

\section{Results}
\label{sec:results}

This section aims to answer the following questions: 
\begin{enumerate}
\item Is \metapg capable of evolving algorithms that improve upon performance, \generalizabilitynospace, and stability in different practical settings? 
\item How well do discovered algorithms do in environments different from those used to evolve them?
\item Are the evolutionary results interpretable?
\end{enumerate}

\begin{figure}[t]
\begin{center}
\includegraphics[width=0.8\textwidth]{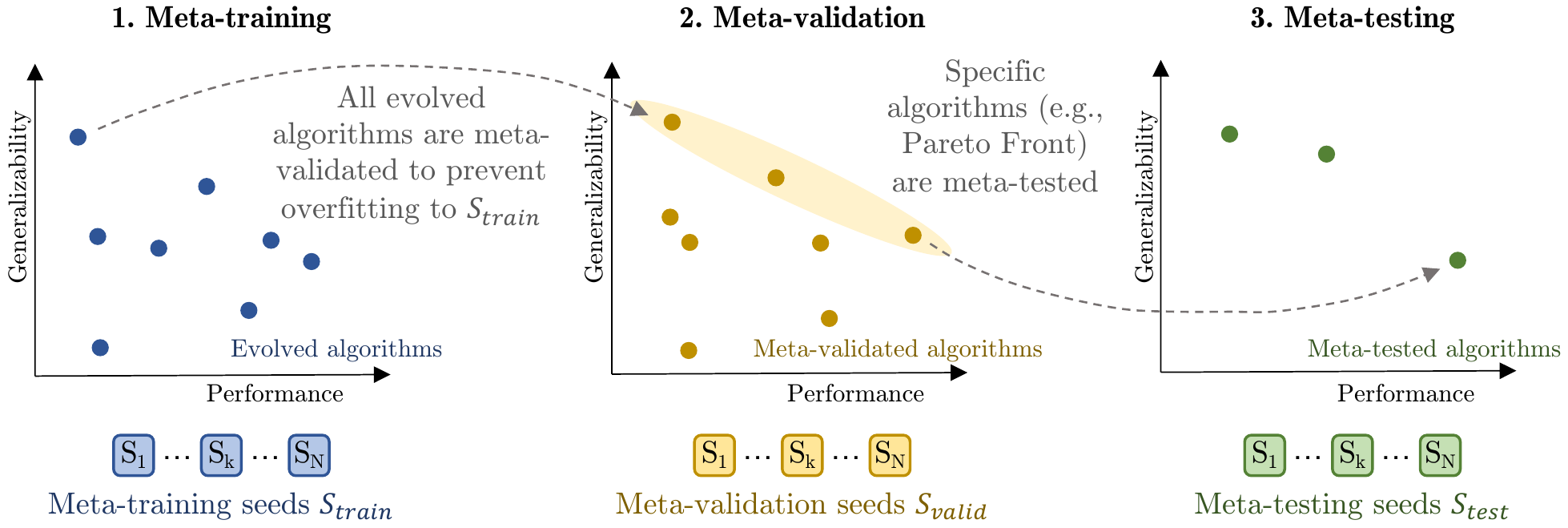}
\end{center}
\caption{\small Running an experiment with \metapg is divided into three phases. \textbf{1. Meta-training}: we evolve a population of algorithms using a set of random seeds $S_{train}$ to compute scores. \textbf{2. Meta-validation}: to prevent overfitting, we reevaluate the scores of all algorithms in the population using a different set of random seeds $S_{valid}$. \textbf{3. Meta-testing}: specific algorithms such as those in the Pareto Front are tested in different environments using a third set of random seeds $S_{test}$.}
\label{fig:train_valid_test}
\end{figure}

We run the experiments following the process represented in Figure \ref{fig:train_valid_test}; we divide them into meta-training, meta-validation, and meta-testing phases. Each phase relies on a different set of $N$ random seeds: $S_{train}$, $S_{valid}$, and $S_{test}$, respectively. Each \metapg run begins with the meta-training phase, which consists of the evolution process described in the previous section. The result of this phase is a population of evolved algorithms, each with a pair of meta-training fitness scores. Since the evolution process is non-deterministic, we run each experiment multiple times without configuration changes and aggregate all resulting populations into one single larger population.

Then, to avoid selecting algorithms that overfit to the set of seeds $S_{train}$, we reevaluate all algorithms in the population with a different set of seeds $S_{valid}$; this corresponds to the meta-validation phase, which provides updated fitness scores for all algorithms. Finally, to assess the fitness of specific algorithms (e.g., the Pareto Front) when deployed in different environments, we use a third set of seeds $S_{test}$ that provides realistic fitness scores in the new environments; this corresponds to the meta-testing phase. In our analyses of the results, we focus on the set of meta-validated algorithms and then meta-test some of them.

\subsection{Training setup}

\textbf{Training environments}\quad We use as training environments: Cartpole and Walker from the RWRL Environment Suite \citep{dulac2021challenges}, Gym Pendulum, and Ant and Humanoid from the Brax physics simulator \citep{brax2021github}. We define different instances of these environments by varying the pole length in Cartpole, the thigh length in Walker, the pendulum length and mass in Pendulum, and, to mimic a practical setting, the mass, friction coefficient, and torque in Ant and Humanoid. See Appendix \ref{sec:env_configs} for the specifics.

\textbf{Meta-training details}\quad The population and maximum graph size consist of 100 individuals and 80 nodes in the Brax environments, respectively, and 1,000 individuals and 60 nodes in the rest of the environments. All are initialized using SAC as a warm-start (see Appendix \ref{sec:warm_start}). For RL algorithm evaluation, we use 10 different seeds $S_{train}$ and fix the number of evaluation episodes $N_{eval}$ to 20. In the case of Brax, since training takes longer, we use 4 different seeds but increase $N_{eval}$ to 32. We meta-train using 100 TPU 1x1 v2 chips for 4 days in the case of Brax environments ($\sim$200K and $\sim$50K evaluated graphs in Ant and Humanoid, respectively), and using 1,000 CPUs for 10 days in the rest of environments ($\sim$100K evaluated graphs per experiment). In all cases we normalize the fitness scores to the range [0, 1]. We set $\kappa = 1$ in \eqref{eq:multi-objective}. Additional details are in Appendix \ref{sec:implementation_details}.

\textbf{Meta-validation details}\quad During meta-validation, we use a set of 10 seeds $S_{valid}$, disjoint with respect to $S_{train}$. In the case of Brax environments, we use 4 meta-validation seeds. Same applies during meta-testing. In each case, we use a number of seeds that achieve a good balance between preventing overfitting and having affordable evaluation time. The value of $N_{eval}$ during meta-validation and meta-testing matches the one used in meta-training.

\textbf{Hyperparameter tuning}\quad We use the same fixed hyperparameters during all meta-training. Algorithms are also meta-validated using the same hyperparameters. In the case of Brax environments, we do hyperparameter-tune the algorithms during meta-validation; additional details can be found in Appendix \ref{sec:hp_tuning_brax}. We also hyperparameter-tune all baselines we compare our evolved algorithms against.

\textbf{RL Training details}\quad The architecture of the policies corresponds to two-layer MLPs with 256 units each. Additional training details are presented in Appendix \ref{sec:training_details}.

\subsection{Optimizing performance, \generalizabilitynospace, and stability in RWRL Environment Suite}

\begin{figure}[tb]
\centering
\begin{subfigure}{.48\textwidth}
  \centering
  \includegraphics[width=.8\linewidth]{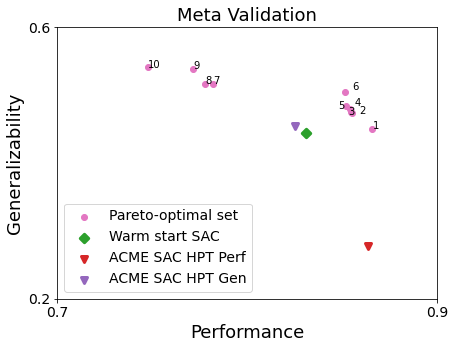}
  \caption{Stability-adjusted fitness scores (computed using Equation \ref{eq:multi-objective}) for algorithms in the Pareto-optimal set.}
  \label{fig:population_cartpole_detail}
\end{subfigure}%
\hspace{1em}%
\begin{subfigure}{.48\textwidth}
  \centering
  \includegraphics[width=0.9\textwidth]{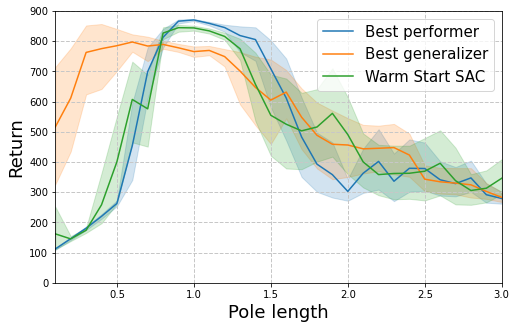}
  \caption{Average return and standard deviation across seeds when evaluating trained policies in multiple RWRL Cartpole instances with different pole lengths (Training configuration is 1.0, see Appendix \ref{sec:env_configs}).}
  \label{fig:perturbation_cartpole}
\end{subfigure}
\caption{\small Evolution results (meta-validation across 10 different seeds) alongside the warm-start algorithm (SAC), and the hyperparameter-tuned ACME SAC when using the RWRL Cartpole environment for training. We show the Pareto Front of algorithms that results after merging the 10 populations corresponding to the 10 repeats of the experiment. The best performer and best generalizer correspond to the algorithms with the highest stability-adjusted performance and \generalizability scores, respectively, according to Equations \ref{eq:perf}, \ref{eq:gen}, and \ref{eq:multi-objective}.}
\label{fig:cartpole_evolution}
\end{figure}

We apply \metapg to RWRL Cartpole and compare the evolved algorithms in the meta-validated Pareto Front with the warm-start SAC and ACME SAC \citep{hoffman2020acme} (Figure \ref{fig:population_cartpole_detail}). When running ACME SAC we first do hyperparameter tuning and pick the two configurations that lead to the best stability-adjusted performance and the best stability-adjusted \generalizability (ACME SAC HPT Perf and ACME SAC HPT Gen, respectively). In Table \ref{tab:comparison_cartpole_detail}, we show numeric values for each of the three metrics. In the case of stability, we show a measure of instability represented as the change in standard deviation compared to the warm-start (i.e., warm-start has the value 1.0). We independently measure instability with respect to performance and \generalizabilitynospace.

The results from Table \ref{tab:comparison_cartpole_detail} show that \metapg discovers RL algorithms that improve upon the warm-start's and ACME SAC's performance, \generalizabilitynospace, and stability in the same environment we used during evolution. Compared to the warm-start, the best performer achieves a 4\% improvement in the stability-adjusted performance score (from 0.836 to 0.868), the best generalizer achieves a 20\% increase in the stability-adjusted \generalizability score (from 0.460 to 0.551), and the selected algorithm in the Pareto-optimal set (Pareto point 6) achieves a 2\% and a 12\% increase in both stability-adjusted performance and stability-adjusted \generalizabilitynospace, respectively. Then, in terms of the stability objective, the best performer reduces performance instability by 67\% and the best generalizer achieves a reduction of 30\% for \generalizability instability.

The gains in \generalizability and stability are substantial when comparing the results to hyperparameter-tuned ACME SAC. The best generalizer achieves a 15\% increase in stability-adjusted \generalizability compared to ACME SAC tuned for such metric. The instability in the hyperparameter-tuned SAC is twice as high (1.48 vs. 0.70, as shown in Table \ref{tab:comparison_cartpole_detail}). In terms of performance, the best performer achieves a slightly better result compared to SAC hyperparameter-tuned for performance.

We report the complete results in Appendix \ref{sec:additional_results_cartpole} and repeat the same experiments in RWRL Walker (Appendix \ref{sec:additional_results_walker}) and Gym Pendulum (Appendix \ref{sec:additional_results_pendulum}) and observe that \metapg also discovers a Pareto Front of algorithms that outperform SAC in both environments. Additional information on the stability of the algorithms is presented in Appendix \ref{sec:stability_analyses}.

Figure \ref{fig:perturbation_cartpole} compares how the best performer and the best generalizer behave in different instances of the environment in which we change the pole length (all instances form the environment set $\mathcal{E}$ used during evolution). We follow the same procedure described by \cite{dulac2021challenges}. The best performer achieves better return in the training configuration than the warm-start's. The best generalizer in turn achieves a lower return but it trades it for higher returns in configurations outside of the training regime, being better at zero-shot generalization. The same behavior holds when using RWRL Walker and Gym Pendulum as training environments (see Appendix \ref{sec:additional_results_walker} and \ref{sec:additional_results_pendulum}, respectively).


\begin{table}[!tb]
\def\arraystretch{1.2}
\begin{center}
\caption{We compare three algorithms in the Pareto Front with SAC (warm-start and hyperparameter-tuned ACME SAC) using metrics obtained in the RWRL Cartpole environment: average performance and \generalizabilitynospace, stability-adjusted performance and \generalizability scores, and measure of instability (standard deviation $\sigma$ divided by the warm-start's $\sigma_{WS}$). We compute these metrics across 10 seeds and the best result in a \emph{column} is \textbf{bolded}. $^{\dagger}$In contrast to performance and \generalizabilitynospace, the lower the instability the better.}
\label{tab:comparison_cartpole_detail}
\begin{tabular}{lcc|cc|cc}
                                           & \multicolumn{2}{c|}{\textbf{Performance}} & \multicolumn{2}{c|}{\textbf{\Generalizabilitynospace} } & \multicolumn{2}{c}{\textbf{Instability}$^{\dagger}$ ($\sigma$/$\sigma_{WS}$)} \\ \cline{2-7} 
\textbf{RL Algorithm}                      & $f_{perf}$       & $\tilde{f}_{perf}$       & $f_{gen}$           & $\tilde{f}_{gen}$          & Perf.              & Gen.             \\ \hline
\textbf{Pareto point 1: Best performer} & \textbf{0.871} & \textbf{0.868} & 0.475 & 0.459 & 0.33 & \textbf{0.59} \\
\textbf{Pareto point 6} & 0.854 & 0.852 & 0.531 & 0.514 & 0.22 & 0.63 \\
\textbf{Pareto point 10: Best generalizer} & 0.770 & 0.756 & \textbf{0.570} & \textbf{0.551} & 1.55 & 0.70 \\ \hline
\textbf{Warm-start SAC} & 0.845 & 0.836 & 0.487 & 0.460 & \textit{1.0} & \textit{1.0} \\
\textbf{ACME SAC HPT Perf} & 0.865 & 0.864 & 0.372 & 0.312 & \textbf{0.11} & 2.22 \\
\textbf{ACME SAC HPT Gen} & 0.845 & 0.833 & 0.518 & 0.478 & 1.33 & 1.48               \\ \hline
\end{tabular}
\end{center}
\end{table}

\subsection{Transferring evolved algorithms between Brax environments}

\begin{figure}[!h]
\begin{center}
\includegraphics[width=0.99\textwidth]{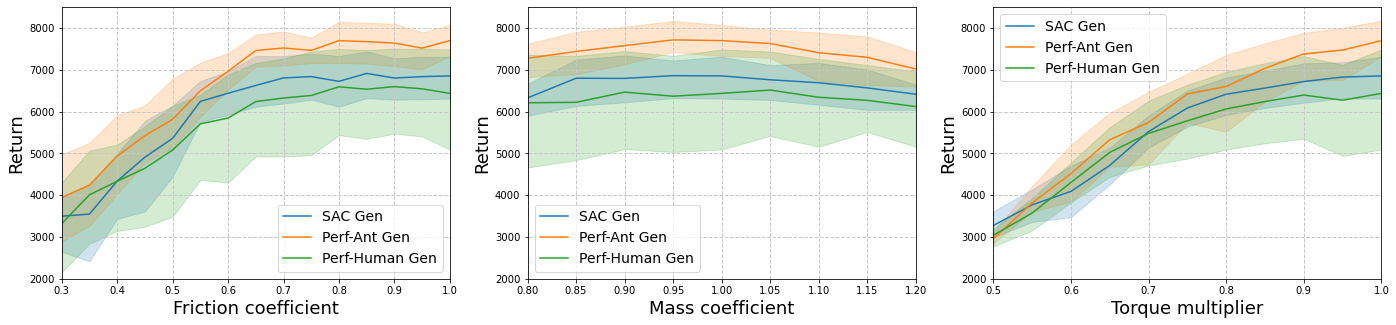}
\end{center}
\caption{Meta-testing in Brax Ant. We compare, after hyperparameter tuning, loss functions evolved in Brax Ant, Brax Humanoid (to assess cross-domain transfer), and the SAC baseline used as warm-start. We plot the average return and standard deviation across random seeds when evaluating in multiple Brax Ant instances with different friction coefficients, mass coefficients, and torque multipliers (in all cases 1.0 is used as training configuration). The loss function evolved in Ant improves upon SAC's performance and generalizability.}
\label{fig:perturbation_ant}
\end{figure}

Figure \ref{fig:perturbation_ant} shows the behaviour of evolved algorithms when meta-tested in perturbed Brax Ant environments (changes in friction coefficient, mass, and torque, see Appendix \ref{sec:env_configs}). We first evolve algorithms independently in both Ant and Humanoid, then select those algorithms that have the highest stability-adjusted \generalizability score $\tilde{f}_{gen}$ during meta-validation. We then re-evaluate using the meta-testing seeds. We compare algorithms evolved in Ant and Humanoid with hyperparameter-tuned SAC.

These results highlight that an algorithm evolved by \metapg in Brax Ant performs and generalizes better than a SAC baseline. Specifically, we observe a 15\% improvement in stability-adjusted performance and 10\% improvement in stability-adjusted \generalizabilitynospace. We also obtain a 23\% reduction in instability. In addition, we observe that an algorithm initially evolved using Brax Humanoid and meta-validated in Ant transfers reasonably well to Ant during meta-testing, achieving slight loss of performance compared to hyperparameter-tuned SAC (adjusting for stability, 17\% less performance and 13\% less \generalizability compared to SAC). We evolved fewer graphs in the case of Humanoid (50K compared to 200K evolved graphs for Brax Ant), as training a policy in Humanoid is more costly. We expect these results to improve if more algorithms are evolved in the population. We present complete results for the Brax Ant environment in Appendix \ref{sec:additional_results_ant} and repeat the same analysis for the Brax Humanoid environment in Appendix \ref{sec:additional_results_humanoid}, in which we observe similar results.

\subsection{Analyzing the evolved RL algorithms}

Next, we analyze evolved algorithms from our experiments on RWRL Cartpole. We pick the best meta-validated performer and generalizer, both evolved from the warm-start SAC (see Appendix \ref{sec:warm_start} for its graph representation in the search space). The policy loss $L_{\pi}$ and critic losses $L_{Q_i}$ (one for each critic network $Q_i$, see Appendix \ref{sec:nodes_list} for details) observed from the graph structure for the best performer are the following:
\begin{align}
    L_{\pi}^{perf} &= \mathbb{E}_{(s_t, a_t, s_{t+1})\sim \mathcal{D}} \left[ \log(\textcolor{blue}{\min(}\pi(\textcolor{blue}{\tilde{a}_{t+1}}|\textcolor{blue}{{s_{t+1}}})\textcolor{blue}{, \gamma)}) - \underset{i}{\min}\,Q_i(s_t, \tilde{a}_t)\right] \\
    L_{Q_i}^{perf} &= \mathbb{E}_{(s_t, a_t, r_t, s_{t+1})\sim \mathcal{D}} \left[ \left(r_t + \gamma\left(\textcolor{red}{\underset{i}{\min}\,}Q_{targ_i}(s_{t+1}, \tilde{a}_{t+1}) \textcolor{red}{- \log\pi(\tilde{a}_{t+1}|s_{t+1})} \right) - Q_i(s_t, a_t)\right)^2\right]
\end{align}
where $\tilde{a}_t \sim \pi(\cdot|s_t)$, $\tilde{a}_{t+1} \sim \pi(\cdot|s_{t+1})$, and $\mathcal{D}$ is an experience dataset extracted from the replay buffer. We highlight in \textcolor{blue}{blue} the changes and additions with respect to SAC, and in \textcolor{red}{red} the elements of the SAC loss function that evolution removes. The loss equations for the best generalizer are:
\begin{align}
    L_{\pi}^{gen} &= \mathbb{E}_{(s_t, a_t,s_{t+1})\sim \mathcal{D}} \left[ \log\pi(\tilde{a}_t|s_t) - \underset{i}{\min}\,Q_i(\textcolor{blue}{s_{t+1}}, \tilde{a}_t)\right] \label{eq:cartpole-gen-policy-loss}  \\
    L_{Q_i}^{gen} &= \mathbb{E}_{(s_t, a_t, r_t, s_{t+1})\sim \mathcal{D}} \left[ \textcolor{blue}{\mathrm{atan}}\left(\left(r_t + \gamma\left(\underset{i}{\min}\,Q_{targ_i}(s_{t+1}, \textcolor{blue}{\tilde{a}_t}) - \log\pi(\textcolor{blue}{\tilde{a}_t}|\textcolor{blue}{s_t})\right)- Q_i(s_t, a_t)\right)^2\right)\right]
\end{align}
\begin{figure}[!h]
\centering
\begin{subfigure}{.48\textwidth}
  \centering
  \includegraphics[width=0.7\textwidth]{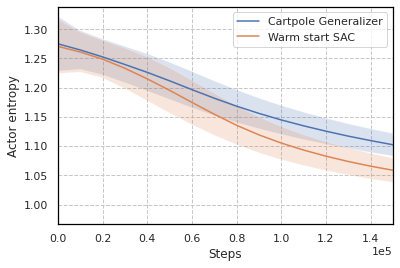}
  \caption{Average entropy of the policy during training for RWRL Cartpole.}
  \label{fig:cartpole_entropy_150000}
\end{subfigure}%
\hspace{1em}%
\begin{subfigure}{.48\textwidth}
  \centering
  \includegraphics[width=0.7\linewidth]{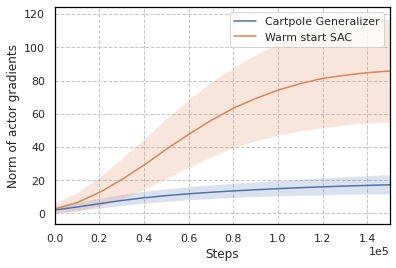}
  \caption{Average gradient norm of the actor loss during training for RWRL Cartpole.}
  \label{fig:cartpole_grad_norm_150000}
\end{subfigure}
\caption{\small Analysis of the entropy and gradient norm of the actor when evaluating the best generalizer from RWRL Cartpole in comparison to the warm-start. We hypothesize this increase in entropy and decrease in gradient norm with respect to SAC contribute to achieve better \generalizabilitynospace.} 
\label{fig:cartpole_entropy_gradnorm_150000}
\end{figure}

While both algorithms resemble the warm-start SAC (see Appendix \ref{sec:warm_start}), we observe that the best performer does not include the entropy term in the critic loss while the best generalizer does (i.e., they correspond to setting $\alpha$ to 0 and 1 in the original SAC algorithm \citep{haarnoja2018soft}, respectively). This aligns with the hypothesis that, since ignoring the entropy pushes the agent to exploit more and explore less, the policy of the best performer overfits better to the training configuration compared to SAC. In contrast, the best generalizer is able to explore more. Figure \ref{fig:cartpole_entropy_150000} validates the latter observation showing a higher entropy for the best generalizer's actor compared to the warm-start's.

The use of arctangent in the critic loss of the best generalizer is also noticeable as, supported by Figure \ref{fig:cartpole_grad_norm_150000}, we observe this operation serves as a way of clipping the loss, which makes gradients smaller and thus prevents the policy's parameters from changing too abruptly. In our experiments, we fix the number of training episodes as a compromise between achievable returns and evaluation runtimes. Clipping the loss has then an early-stopping effect compared to the baseline and results in a policy less overfitted, which benefits generalization. In Appendix \ref{sec:extended_analysis_cartpole}, we show both extended results that ignore the fix budget requirement and the equations for the best evolved algorithms in the remaining environments. Appendix \ref{sec:eqs_walker_pendulum} and \ref{sec:eqs_ant_humanoid} present the equations for the other RWRL environments and the Brax environments, respectively.

\subsection{Discussion \& Future Work}

We have shown that MetaPG can discover novel RL loss functions that achieve better training stability and zero-shot generalization compared to warm-start algorithms such as SAC. Our search space consists only of primitive operators similar to \citet{co2021evolving}. While this promotes expressiveness of algorithms in the space, it requires many nodes and edges to represent a loss function which makes our search space vast and good loss functions are extremely sparse in this space. It is challenging for evolutionary algorithms to traverse this space given limited search budget. One area for improvement is to design a more efficient search space so that we have a greater chance of discovering better algorithms under the same computation budget. 

Another direction of future work is to improve the transferrability of evolved algorithms to new domains. We have observed that, while evolved algorithms transfer reasonably well (especially best performers, see Appendix \ref{sec:transferring_algorithms}), sometimes they do not perform better than SAC in the new environments without hyperparameter tuning first. At the same time, it poses an interesting research question of determining whether \metapg is better suited to find ``super algorithms'' for specific environments or a new generation of all-purpose algorithms.

Finally, it would be interesting to ensemble the evolved loss functions on the Pareto front. Such loss function may give additional flexibility for practitioners when designing an RL system by encoding complex design choices into an interpolation across objectives. We hope the released dataset of evolved algorithms help the community gain deeper insights on differences between loss functions.

\section{Conclusion}

We presented \metapgnospace, a method that evolves actor-critic RL loss functions to optimize multiple RL objectives simultaneously and applied it to discovering algorithms that perform well, achieve zero-shot generalization across different environment configurations, and are stable; a triad of objectives with real-world implications. The experiments in RWRL Cartpole, RWRL Walker, and Gym Pendulum demonstrated that \metapg discovered algorithms that, when using one environment during evolution and then meta-validating in that same environment, outperform SAC, achieving a 4\% and 20\% improvement in stability-adjusted performance and stability-adjusted \generalizabilitynospace, respectively, and a reduction of up to 67\% in instability. Experiments on Brax Ant and Brax Humanoid proved evolution is successful in more complex environments, achieving a 15\% and 10\% increase in stability-adjusted performance and stability-adjusted \generalizabilitynospace, respectively. We also observed that, when transferring evolved algorithms to environments different from those used during evolution, the loss of performance and \generalizability in the new environment is minimal and is comparable to SAC. Finally, we have analyzed the evolved loss functions and linked specific elements in their structure to fitness results, such as the removal of the entropy term to benefit performance.




\subsubsection*{Author Contributions}
Hidden for the double-blind review process.

\subsubsection*{Acknowledgments}
Hidden for the double-blind review process.

\bibliography{main}
\bibliographystyle{tmlr}

\appendix

\section{Search space details}
\label{sec:search_space}

In this section we present the details of the search space; we divide the nodes into input, output, and operation nodes. Output nodes correspond to losses computed by the algorithm, whose gradient with respect to the algorithm inputs is then computed in a training loop. In this paper we do steepest gradient descent to update network parameters; given the search space complexity, we leave incorporating other gradient descent strategies into the search space out of its scope. However, other strategies such as natural gradient or conjugate gradient could be incorporated, as they do gradient transformations and are agnostic to loss functions.

During the training process using an evolved algorithm, agents then learn the policy by means of experience tuples coming from a replay buffer. \metapg admits both continuous and discrete action spaces; specific nodes in the graphs ---e.g., the neural networks--- are adapted to work with the corresponding space.

During the evolution process, we fix a maximum number of nodes (60 and 80 in our experiments), which consists of the aforementioned input and output nodes, and several operation nodes. The majority of operation nodes treat input elements as tensors with variable shapes in order to maximize graph flexibility. Each node possesses a certain number of input and output edges, which are determined by the specific operation this node carries out. For example, a node that takes in two tensors and multiplies them element-wise has two input edges and a single output edge.

\subsection{List of nodes}
\label{sec:nodes_list}

A complete list of the nodes considered follows:

\paragraph{Input nodes} We only encode canonical RL elements as inputs:
\begin{itemize}\setlength{\itemsep}{0pt}
    \item Policy network $\pi$
    \item Two critic networks, $Q_1$ and $Q_2$, and two target critic networks, $Q_{targ_1}$ and $Q_{targ_2}$
    \item Batch of states $s_t$ and next states $s_{t+1}$
    \item Batch of actions $a_t$
    \item Batch of rewards $r_t$
    \item Discount factor $\gamma$
\end{itemize}

\paragraph{Output nodes} The output of these nodes is used as loss function to compute gradient descent on:
\begin{itemize}\setlength{\itemsep}{0pt}
    \item Policy loss $L_{\pi}$
    \item Critic loss $L_{Q_i}$
\end{itemize}

\paragraph{Operation nodes} These nodes operate generally on tensors and can broadcast operations when input sizes do not match: 
\begin{itemize}\setlength{\itemsep}{0pt}
    \item Addition: add two, three, or four tensors
    \item Multiplication: compute element-wise product of two or three tensors
    \item Subtract two tensors
    \item Divide two tensors and add constant $\epsilon$ to the denominator
    \item Neural network operations: Action distribtuion from state, stopping gradient computation
    \item Operations with action distributions: Sample, Log-probability
    \item Mean, sum, and standard deviation over last axis of array or over entire array
    \item Cumulative sum, cumulative sum with discount
    \item Squared difference
    \item Multiply by a constant: -1, 0.1, 0.01, 0.5, 2.0
    \item Minimum and maximum over last axis of a tensor
    \item Minimum and maximum element-wise between two tensors
    \item Other general operations: clamp, absolute value, square, logarithm, exponential
    \item Trigonometry functions
\end{itemize}

\subsection{Size of the search space}
\label{sec:size_of_search_space}

To get an upper bound of the size of the search space in terms of the number of possible graphs (not all of them valid), we consider the ($k$+1)-th node in the graph of size $K$ nodes. This node can correspond to one of the $N$ different operation nodes. Assuming that this node has two inputs, there are $\binom{k}{2}$ possibilities of connecting to previous nodes in the graph. Therefore, we have a total of $\frac{1}{2}Nk(k-1)$ possible combinations for the ($k$+1)-th node. Then, an upper bound of the total number of possible graphs is
\begin{equation}
    \left(\frac{NK(K-1)}{2}\right)^K
\end{equation}
With $N=33$ and $K=60$ we obtain approximately 10$^{286}$ graphs. This number increases to 10$^{401}$ if $K=80$ instead.

\section{Additional implementation details}
\label{sec:implementation_details}

\paragraph{Multi-objective evolution}
We do 10 independent repeats of \metapg when evolving on RWRL Cartpole, RWRL Walker, and Gym Pendulum; 5 repeats when evolving on Brax Ant; and 3 repeats for Brax Humanoid.


\paragraph{Warm-starting} Algorithms are initialized using the warm-start SAC graph (see Appendix \ref{sec:warm_start}), which consists of 33 nodes. Additional operation nodes are added to each individual until reaching the maximum amount of 60 nodes.

\paragraph{Mutation} During mutation, there is a 50\% chance an individual undergoes node mutation and a 50\% chance it undergoes edge mutation. During node mutation, there is a 50\% chance of replacing one node, a 25\% chance of replacing 2 nodes, a 12.5\% chance of replacing 4 nodes, and a 6.25\% chance of replacing 8 and 16 nodes, respectively. During edge mutation, only one edge in the graph is replaced randomly.

\paragraph{Hashing} In the hashing process we use a fixed set of synthetic inputs with a batch size of 16.

\paragraph{Encoding multiple objectives} \metapg keeps the population to a fixed size during evolution. To decide which individuals should be removed in the process, the method makes use of different fitness scores that encode each of the RL objectives considered and are adjusted for stability. These scores are not combined but treated separately in a multi-objective fashion. This means that, after evaluating a graph $i$, it will have fitness scores $\{f_{i,1}, f_{i,2}, ..., f_{i,F}\}$, where $F$ is the number of objectives considered. Then, when comparing two graphs $i$ and $j$, we say $i$ has higher fitness than $j$ iff $f_{i,k} \geq f_{j,k}, \forall k$, with at least one fitness score $k'$ such that $f_{i,k'} > f_{j,k'}$. In this case we also say graph $i$ Pareto-dominates graph $j$. If neither $i$ Pareto-dominates $j$ nor vice versa, we say both graphs are Pareto-optimal (e.g., if $f_{i,k'} > f_{j,k'}$ and $f_{i,k''} < f_{j,k''}$, then $i$ and $j$ are Pareto-optimal).

The process of removing individuals from the population follows the NSGA-II algorithm \citep{deb2002fast} which, assuming a maximum population size of $P_{max}$ individuals:
\begin{enumerate}
    \item From a set of $P$ individuals, with $|P| > P_{max}$, it computes the set $P_{opt}$ of Pareto-optimal fittest graphs. None of the graphs in $P_{opt}$ is Pareto-dominated by any other graph in the population and, if a graph $i$ in $P$ is Pareto-dominated by at least one other graph, then $i$ does not belong to the Pareto-optimal set.
    \item If $|P_{opt}| \geq P_{max}$, the graphs are ranked based on their crowding distance in the fitness space. This favors individuals that are further apart from other individuals in the fitness space. The fittest $P_{max}$ individuals of the Pareto-optimal set $P_{opt}$ are kept in the population.
    \item Otherwise, if $|P_{opt}| < P_{max}$, the set $P_{opt}$ is kept in the population and the process is repeated taking $P \leftarrow P$ \textbackslash $P_{opt}$ and $P_{max} \leftarrow P_{max} - |P_{opt}|$.
\end{enumerate}


\section{Environment Configurations}
\label{sec:env_configs}

In this work we use multiple environments for our experiments: Cartpole and Walker from the RWRL Environment Suite \citep{dulac2021challenges}, Gym Pendulum, and Ant and Humanoid from the Brax physics simulator \citep{brax2021github}. In Table \ref{tab:environment_configurations} we list the training configuration used for each and the parameters that we use to assess the \generalizability of the policies. The parameters that are not listed are fixed to the default values for the environment in question.

In the case of Gym Pendulum and the Brax environments, we have more than one perturbation parameter; the \generalizability score is computed by first sweeping through the perturbation values for one, taking the average $f_{gen_1}$, repeating the same process for the others to compute $f_{gen_2}$, ..., $f_{gen_P}$, and then computing the average of both to get the final score, i.e., $f_{gen} = (f_{gen_1} + ... + f_{gen_P}) / P$, where $P$ is the total number of different parameters that are perturbed.

For the parameters that undergo perturbations, we select the specific value used in the training configuration based on the following criteria: in the case of the RWRL Environment Suite, we pick the training value used by \cite{dulac2021challenges}, which corresponds to the default value as described in each benchmark; in the case of Gym Pendulum, we pick the default values given by the environment as training configuration; in the case of Brax, we follow the rationale of selecting training configurations that represent a baseline scenario (i.e., no mass variations, normal friction, and normal torque).

\begin{table}[!h]
\begin{center}
\caption{Environment parameters and perturbations.}
\label{tab:environment_configurations}
\begin{tabular}{lc}
\hline 
                 \cellcolor[HTML]{DFDFDF}            \textbf{Environment parameter}       &\cellcolor[HTML]{DFDFDF} \textbf{Value}     \\ \hline
                            
                            \multicolumn{2}{c}{\cellcolor[HTML]{EFEFEF} \textbf{RWRL Cartpole}} \\ \hline
                            
\textbf{Rollout length} &  1,000   \\
\textbf{Min. return} & 0  \\
\textbf{Max. return} & 1,000  \\
\textbf{Training episodes} & 150 \\ 
\textbf{Perturbation parameter (PP)} & Pole length  \\
\textbf{PP Default value} & 1.0  \\
\textbf{PP \Generalizability values} & 0.1 to 3.0 in steps of 0.1 \\ \hline
\multicolumn{2}{c}{\cellcolor[HTML]{EFEFEF} \textbf{RWRL Walker}} \\ \hline
                            
\textbf{Rollout length} &  1,000   \\
\textbf{Min. return} & 0  \\
\textbf{Max. return} & 1,000  \\
\textbf{Training episodes} & 225 \\ 
\textbf{Perturbation parameter (PP)} & Thigh length  \\
\textbf{PP Default value} & 0.225  \\
\textbf{PP \Generalizability values} & .1, .125, .15, .175, .2, .225, .25, .3, .35, .4, .45, .5, .55, .6, .7 \\ \hline
\multicolumn{2}{c}{\cellcolor[HTML]{EFEFEF} \textbf{Gym Pendulum}} \\ \hline
                            
\textbf{Rollout length} &  2,000   \\
\textbf{Min. return} & -2,000  \\
\textbf{Max. return} & 0  \\
\textbf{Training episodes} & 100 \\ 
\textbf{Perturbation parameter 1 (PP1)} & Pendulum mass  \\
\textbf{PP1 Default value} & 1.0  \\
\textbf{PP1 \Generalizability values} & .1, .2, .4, .5, .75, 1.0, 1.5, 2.0, 3.0, 5.0, 7.5, 10.0 \\
\textbf{Perturbation parameter 2 (PP2)} & Pendulum length  \\
\textbf{PP2 Default value} & 1.0  \\
\textbf{PP2 \Generalizability values} & .1, .2, .4, .5, .75, 1.0, 1.5, 2.0, 3.0, 5.0, 7.5, 10.0 \\ \hline

\multicolumn{2}{c}{\cellcolor[HTML]{EFEFEF} \textbf{Brax environments}} \\ \hline

\textbf{Rollout length} &  1,000 \\
\textbf{Min. return} & 0 \\
\textbf{Max. return} & 10,000 (Ant) and 14,000 (Humanoid)  \\
\textbf{Training episodes} & 1,000 \\ 
\textbf{Perturbation parameter 1 (PP1)} & Mass coefficient  \\
\textbf{PP1 Default value} & 1.0  \\
\textbf{PP1 \Generalizability values} & 0.8 to 1.2 in steps of 0.05 \\
\textbf{Perturbation parameter 2 (PP2)} & Friction coefficient  \\
\textbf{PP2 Default value} & 1.0  \\
\textbf{PP2 \Generalizability values} & 0.3 to 1.0 in steps of 0.05 \\
\textbf{Perturbation parameter 3 (PP3)} & Torque multiplier  \\
\textbf{PP3 Default value} & 1.0  \\
\textbf{PP3 \Generalizability values} & 0.5 to 1.0 in steps of 0.05 \\
\textbf{Perturbation parameter 4 (PP4)} & Combined parameters PP1, PP2, and PP3  \\
\textbf{PP4 Default value} & 1.0 for each  \\
\textbf{PP4 \Generalizability values} & Grid search over individual \generalizability values \\

\hline
\end{tabular}
\end{center}
\end{table}

\section{Warm-start SAC}
\label{sec:warm_start}

\begin{figure}[!h]
\begin{center}
\includegraphics[width=0.99\textwidth]{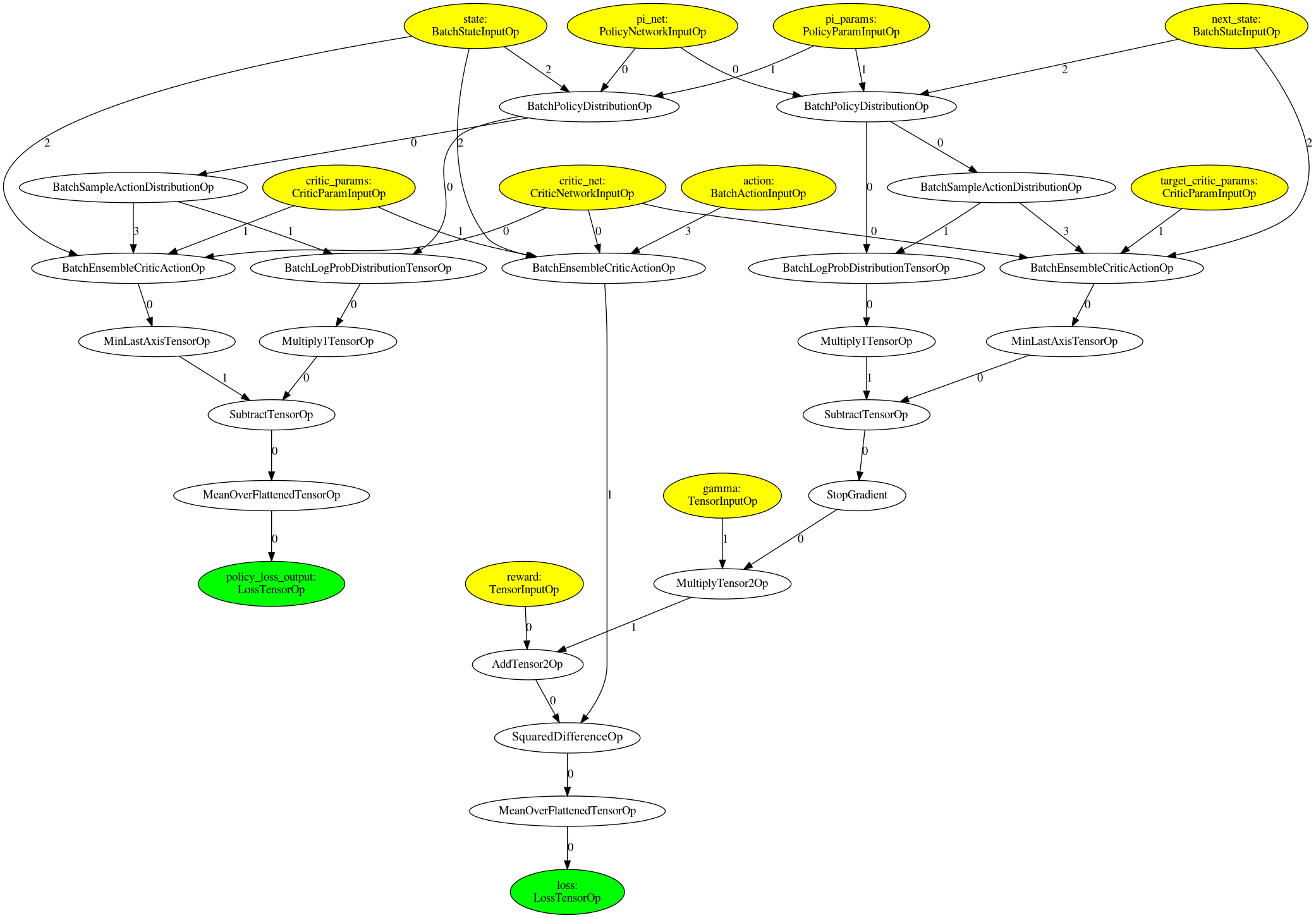}
\end{center}
\caption{Soft Actor-Critic (SAC) algorithm represented as a graph to initialize the population as a warm-start algorithm.}
\label{fig:warm_start}
\end{figure}

We present the version of Soft Actor-Critic (SAC) \citep{haarnoja2018soft} used in this work as the warm-start algorithm to initialize the population. We first present the equations for the policy loss $L_{\pi}^{WS}$ and critic loss $L_{Q_i}^{WS}$:
\begin{align}
    L_{\pi}^{WS} &= \mathbb{E}_{(s_t, a_t)\sim \mathcal{D}} \left[ \log\pi(\tilde{a}_t|s_t) - \underset{i}{\min}\,Q_i(s_t, \tilde{a}_t)\right] \\
    L_{Q_i}^{WS} &=  \mathbb{E}_{(s_t, a_t, r_t, s_{t+1})\sim \mathcal{D}} \left[\left(r_t + \gamma\left(\underset{i}{\min}\,Q_{targ_i}(s_{t+1}, \tilde{a}_{t+1}) - \log\pi(\tilde{a}_{t+1}|s_{t+1})\right) - Q_i(s_t, a_t)\right)^2\right]
\end{align}
where $\tilde{a}_t \sim \pi(\cdot|s_t)$, $\tilde{a}_{t+1} \sim \pi(\cdot|s_{t+1})$, and  $\mathcal{D}$ is a dataset from the replay buffer. Then, in Figure \ref{fig:warm_start} we represent these two equations that define the SAC algorithm in the form of a graph with typed input and outputs. \metapg then modifies this graphs following the procedure described in Section \ref{sec:methods}.

\section{Additional RL training details}
\label{sec:training_details}

An individual encoding a RL algorithm in the form of a graph is evaluated by training an agent using such algorithm. We use an implementation based on an ACME agent \citep{hoffman2020acme} for the RWRL and Gym environments, and an implementation based on the Brax physics simulator \citep{brax2021github} for the Brax environments. The configuration of the training setup are shown in Table \ref{tab:rl_training_details_rwrl} for RWRL and Gym environments, and in Table \ref{tab:rl_training_details_brax} for the Brax environments.

\begin{table}[!h]
\begin{center}
\caption{RL Training setup for the RWRL and Gym environments.}
\label{tab:rl_training_details_rwrl}
\begin{tabular}{c|c} \hline
 
                            \textbf{Parameter}       &\textbf{Value}     \\ \hline

\textbf{Discount factor} $\gamma$ &  0.99 \\
\textbf{Batch size} & 64 (RWRL Cartpole and Gym Pendulum) \\
 & 128 (RWRL Walker) \\
\textbf{Learning rate} & $3\cdot10^{-4}$  \\
\textbf{Target smoothing coeff.} $\tau$ & 0.005 \\ 
\textbf{Replay buffer size} & 1,000,000  \\
\textbf{Min. num. samples in the buffer} & 10,000  \\
\textbf{Gradient updates per learning step} &  1 \\
\textbf{n step} & 1 \\
\textbf{Reward scale} & 5.0 \\ \hline
\textbf{Actor network} & MLP (256, 256) \\
\textbf{Actor activation function} & ReLU \\
\textbf{Tanh on output of actor network} & Yes \\
\textbf{Critic networks} & MLP (256, 256) \\
\textbf{Critic activation function} & ReLU \\ \hline

\end{tabular}
\end{center}
\end{table}

\begin{table}[!h]
\begin{center}
\caption{RL Training setup for the Brax environments.}
\label{tab:rl_training_details_brax}
\begin{tabular}{c|c} \hline
 
                            \textbf{Parameter}       &\textbf{Value}     \\ \hline

\textbf{Discount factor} $\gamma$ &  0.95  \\
\textbf{Batch size} & 128 \\
\textbf{Learning rate} & $6\cdot10^{-4}$ \\
\textbf{Target smoothing coeff.} $\tau$ & 0.005 \\ 
\textbf{Replay buffer size} & 1,000,000 \\
\textbf{Min. num. samples in the buffer} & 1,000 \\
\textbf{Gradient updates per learning step} &  64 \\
\textbf{Reward scale} & 10.0 \\ 
\textbf{Number of parallel environments} & 128 \\ \hline
\textbf{Actor network} & MLP (256, 256) \\
\textbf{Actor activation function} & ReLU \\
\textbf{Tanh on output of actor network} & Yes \\
\textbf{Critic networks} & MLP (256, 256) \\
\textbf{Critic activation function} & ReLU \\ \hline

\end{tabular}
\end{center}
\end{table}

\section{Additional evolution results}
\label{sec:additional_results}

In this section we present the additional evolution results of the paper. We first introduce the remaining figures for RWRL Cartpole, then outline evolution results for RWRL Walker and Gym Pendulum, then show how different algorithms in the population for all three environments compare in terms of stability, and finally provide results for the Brax environments.

\subsection{Evolution results for RWRL Cartpole}
\label{sec:additional_results_cartpole}

\begin{figure}[!h]
\begin{center}
\includegraphics[width=0.8\textwidth]{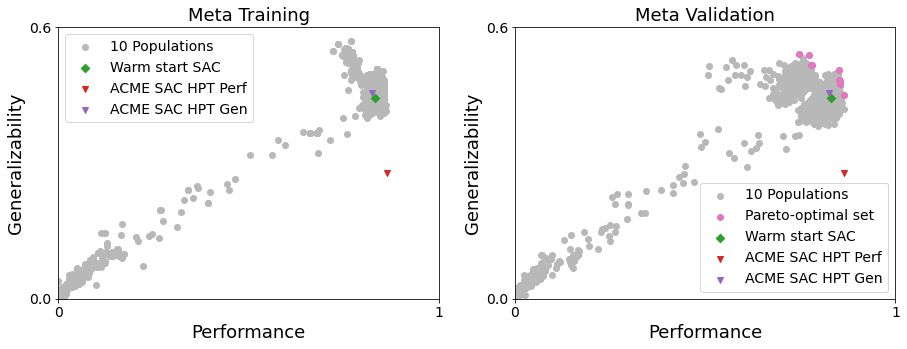}
\end{center}
\caption{Meta-training and meta-validation stability-adjusted fitness scores (computed using Equation \ref{eq:multi-objective} across 10 seeds) for each RL algorithm in the population alongside the warm-start (SAC) and ACME SAC when using the RWRL Cartpole environment for training. We show the meta-validated Pareto Front of algorithms that results after merging the 10 populations corresponding to the 10 repeats of the experiment.}
\label{fig:population_cartpole}
\end{figure}

\begin{figure}[!h]
\begin{center}
\includegraphics[width=0.6\textwidth]{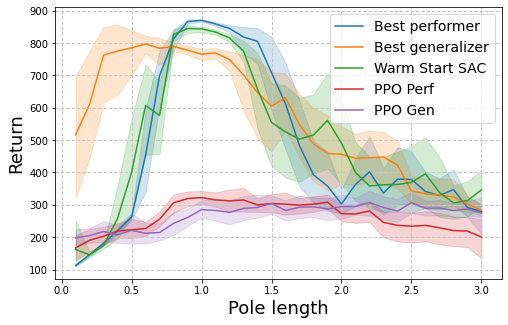}
\end{center}
\caption{Average and standard deviation across seeds of the meta-validation performance when training on a single configuration of RWRL Cartpole and evaluating on multiple unseen ones. We compare the best performer, the best generalizer, the warm start SAC, and we also add two hyperparameter-tuned PPO runs, one tuned for performance and the other for \generalizability (see Appendix \ref{sec:transfer_procedure} for hyperparameter tuning details). The pole length changes across environment configurations and a length of 1.0 is used as training configuration.}
\label{fig:ppo_curves}
\end{figure}

Figure \ref{fig:population_cartpole} shows the resulting population when running evolution using the RWRL Cartpole environment \citep{dulac2021challenges} and Table \ref{tab:comparison_cartpole} shows the raw and stability-adjusted fitness scores for each algorithm in the Pareto Front. Figure \ref{fig:ppo_curves} shows the performance of the evolved algorithms across different environment configurations, it updates Figure \ref{fig:perturbation_cartpole} by introducing PPO \cite{schulman2017proximal} in the comparison. We found that PPO was not well-suited for the continuous control tasks explored in this work.

\begin{table}[!h]
\def\arraystretch{1.2}
\begin{center}
\caption{Similar to Table \ref{tab:comparison_cartpole_detail}, we compare all algorithms in the Pareto Front with SAC (warm-start and hyperparameter-tuned ACME SAC) using metrics obtained in the \textbf{RWRL Cartpole} environment: average performance and \generalizabilitynospace, stability-adjusted performance and \generalizability scores, and measure of instability (standard deviation $\sigma$ divided by the warm-start's $\sigma_{WS}$. We compute these metrics across 10 seeds and the best result in a \emph{column} is \textbf{bolded}. $^{\dagger}$In contrast to performance and \generalizabilitynospace, the lower the instability the better.
}
\label{tab:comparison_cartpole}
\begin{tabular}{lcc|cc|cc}
                                           & \multicolumn{2}{c|}{\textbf{Performance}} & \multicolumn{2}{c|}{\textbf{\Generalizabilitynospace} } & \multicolumn{2}{c}{\textbf{Instability}$^{\dagger}$ ($\sigma$/$\sigma_{WS}$)} \\ \cline{2-7} 
\textbf{RL Algorithm}                      & $f_{perf}$       & $\tilde{f}_{perf}$       & $f_{gen}$           & $\tilde{f}_{gen}$          & Perf.              & Gen.             \\ \hline
\textbf{Pareto 1: Best performer} & \textbf{0.871} & \textbf{0.868} & 0.475 & 0.459 & 0.33 & \textbf{0.59} \\
\textbf{Pareto 2} & 0.857 & 0.856 & 0.513 & 0.488 & \textbf{0.11} & 0.93 \\
\textbf{Pareto 3} & 0.857 & 0.855 & 0.514 & 0.489 & 0.22 & 0.93 \\
\textbf{Pareto 4} & 0.856 & 0.854 & 0.517 & 0.493 & 0.22 & 0.89 \\
\textbf{Pareto 5} & 0.855 & 0.853 & 0.520 & 0.497 & 0.22 & 0.85 \\
\textbf{Pareto 6} & 0.854 & 0.852 & 0.531 & 0.514 & 0.22 & 0.63 \\
\textbf{Pareto 7} & 0.798 & 0.788 & 0.540 & 0.524 & 1.11 & \textbf{0.59} \\
\textbf{Pareto 8} & 0.794 & 0.784 & 0.546 & 0.528 & 1.11 & 0.67 \\
\textbf{Pareto 9} & 0.783 & 0.776 & 0.569 & 0.543 & 0.78 & 0.96 \\
\textbf{Pareto 10: Best generalizer} & 0.770 & 0.756 & \textbf{0.570} & \textbf{0.551} & 1.55 & 0.70 \\ \hline
\textbf{Warm-start SAC} & 0.845 & 0.836 & 0.487 & 0.460 & \textit{1.0} & \textit{1.0} \\
\textbf{ACME SAC HPT Perf} & 0.865 & 0.864 & 0.372 & 0.312 & \textbf{0.11} & 2.22 \\
\textbf{ACME SAC HPT Gen} & 0.845 & 0.833 & 0.518 & 0.478 & 1.33 & 1.48               \\ \hline
\end{tabular}
\end{center}
\end{table}

\subsection{Evolution results for RWRL Walker}
\label{sec:additional_results_walker}

\begin{figure}[!h]
\begin{center}
\includegraphics[width=0.8\textwidth]{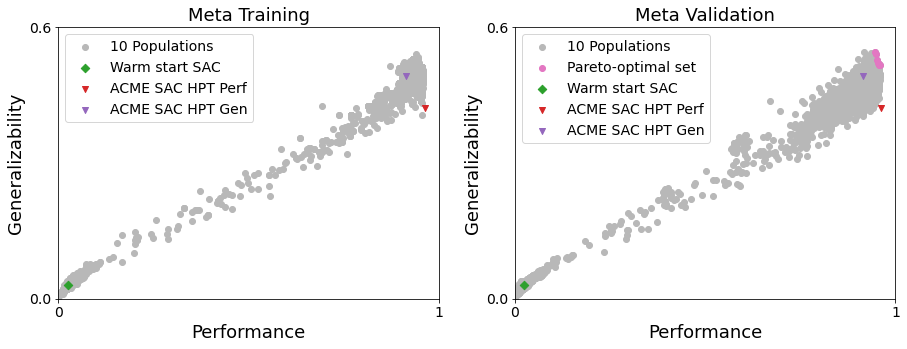}
\end{center}
\caption{Meta-training and meta-validation stability-adjusted fitness scores (computed using Equation \ref{eq:multi-objective} across 10 seeds) for each RL algorithm in the population alongside the warm-start (SAC) and ACME SAC when using the RWRL Walker environment for training. We show the meta-validated Pareto-optimal set of algorithms that results after merging the 10 populations corresponding to the 10 repeats of the experiment.}
\label{fig:population_walker}
\end{figure}

\begin{figure}[!h]
\begin{center}
\includegraphics[width=0.5\textwidth]{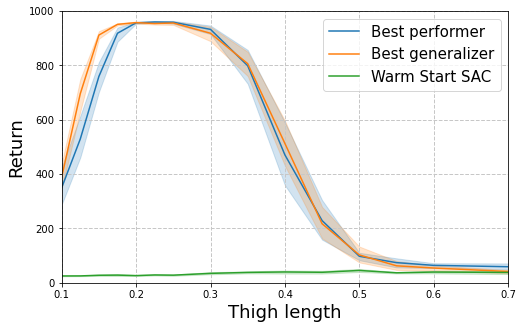}
\end{center}
\caption{Average and standard deviation across seeds of the meta-validation performance of the best performer, the best generalizer, and the warm-start (SAC) when training on a single configuration of RWRL Walker and evaluating on multiple unseen ones. The thigh length changes across environment configurations and a length of 0.225 is used as training configuration.}
\label{fig:perturbation_walker}
\end{figure}

We present evolution results when running \metapg with RWRL Walker as the training environment. In Figures \ref{fig:population_walker} and \ref{fig:perturbation_walker} we show the resulting population and the performance across environment configurations for the best performer and the best generalizer in the Pareto Front, respectively. Exact numbers for each algorithm in the Pareto Front can be found in Table \ref{tab:comparison_walker}. This table also shows the scores of the warm-start and ACME SAC. As covered in Appendix \ref{sec:transfer_procedure}, we do not hyperparameter-tune the warm-start before the experiments. As a result, the warm-start might perform poorly, as is the case in this environment. We can observe \metapg is able to increase the performance and \generalizability of the evolved algorithms during the evolution process.

\begin{table}[!h]
\def\arraystretch{1.2}
\begin{center}
\caption{Similar to Table \ref{tab:comparison_cartpole_detail}, we compare all algorithms in the Pareto Front with SAC (warm-start and hyperparameter-tuned ACME SAC) using metrics obtained in the \textbf{RWRL Walker} environment: average performance and \generalizabilitynospace, stability-adjusted performance and \generalizability scores, and measure of instability (standard deviation $\sigma$ divided by the warm-start's $\sigma_{WS}$. We compute these metrics across 10 seeds and the best result in a \emph{column} is \textbf{bolded}. $^{\dagger}$In contrast to performance and \generalizabilitynospace, the lower the instability the better.
}
\label{tab:comparison_walker}
\begin{tabular}{lcc|cc|cc}
                                           & \multicolumn{2}{c|}{\textbf{Performance}} & \multicolumn{2}{c|}{\textbf{\Generalizabilitynospace} } & \multicolumn{2}{c}{\textbf{Instability}$^{\dagger}$ ($\sigma$/$\sigma_{WS}$)} \\ \cline{2-7} 
\textbf{RL Algorithm}                      & $f_{perf}$       & $\tilde{f}_{perf}$       & $f_{gen}$           & $\tilde{f}_{gen}$          & Perf.              & Gen.             \\ \hline
\textbf{Pareto 1: Best performer} & \textbf{0.963} & \textbf{0.961} & 0.544 & 0.526 & 1.0 & 18.0 \\
\textbf{Pareto 2} & 0.962 & 0.959 & 0.536 & 0.524 & \textbf{1.0} & 12.0 \\
\textbf{Pareto 3} & 0.960 & 0.958 & 0.542 & 0.527 & 1.5 & 15.0 \\
\textbf{Pareto 4} & 0.959 & 0.956 & 0.541 & 0.528 & \textbf{1.0} & 13.0 \\
\textbf{Pareto 5} & 0.960 & 0.955 & 0.541 & 0.532 & 1.5 & 9.0 \\
\textbf{Pareto 6} & 0.954 & 0.951 & 0.555 & 0.546 & 2.5 & 9.0 \\
\textbf{Pareto 7: Best generalizer} & 0.955 & 0.950 & \textbf{0.569} & \textbf{0.554} & 2.5 & 15.0 \\ \hline
\textbf{Warm-start SAC} & 0.028 & 0.026 & 0.033 & 0.032 & \textit{\textbf{1.0}} & \textit{\textbf{1.0}} \\
\textbf{ACME SAC HPT Perf} & 0.968 & 0.965 & 0.444 & 0.430 & 1.5 & 14.0 \\
\textbf{ACME SAC HPT Gen} & 0.926 & 0.918 & 0.510 & 0.498 & 4.0 & 12.0              \\ \hline
\end{tabular}
\end{center}
\end{table}

\subsection{Evolution results for Gym Pendulum}
\label{sec:additional_results_pendulum}

\begin{figure}[!h]
\begin{center}
\includegraphics[width=0.8\textwidth]{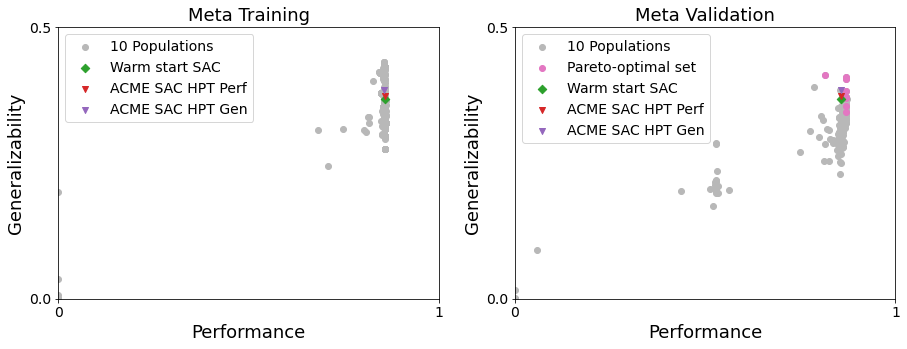}
\end{center}
\caption{Meta-training and meta-validation stability-adjusted fitness scores (computed using Equation \ref{eq:multi-objective} across 10 seeds) for each RL algorithm in the population alongside the warm-start (SAC) and ACME SAC when using the Gym Pendulum environment for training. We show the meta-validated Pareto-optimal set of algorithms that results after merging the 10 populations corresponding to the 10 repeats of the experiment.}
\label{fig:population_pendulum}
\end{figure}

\begin{figure}[!h]
\begin{center}
\includegraphics[width=0.8\textwidth]{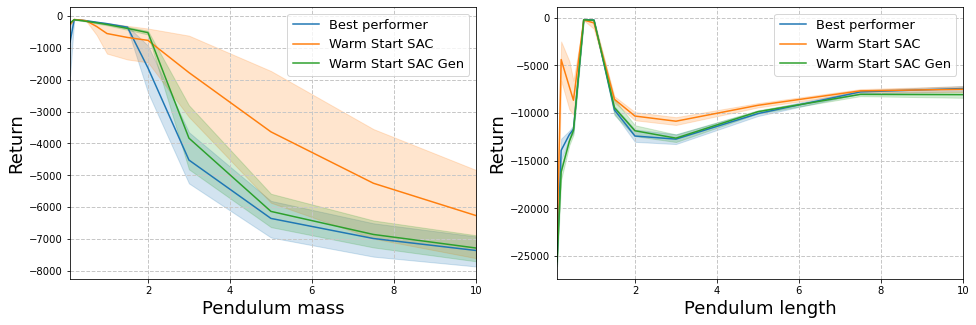}
\end{center}
\caption{Average and standard deviation across seeds of the meta-validation performance of the best performer, the best generalizer, and the warm-start (SAC) when training on a single configuration of Gym Pendulum and evaluating on multiple unseen ones. The pendulum mass and the pendulum length independently change across environment configurations (we change one at a time). The training configurations use a pendulum mass and a pendulum length of 1.0 and 1.0, respectively.}
\label{fig:perturbation_pendulum}
\end{figure}

We present evolution results when running \metapg with Gym Pendulum as the training environment. In Figures \ref{fig:population_pendulum} and \ref{fig:perturbation_pendulum} we show the resulting population and the performance across environment configurations for the best performer and the best generalizer in the Pareto Front, respectively. In the case of Pendulum, the \generalizability fitness score is computed across the perturbation of two different parameters: the pendulum mass and the pendulum length. These parameters are changed separately, as opposed to varying both the mass and length of the pendulum in the same run. Exact numbers can be found in Table \ref{tab:comparison_pendulum}, in which an average improvement over the warm-start of 2\% in performance and 15\% in \generalizability is achieved.

\begin{table}[!h]
\def\arraystretch{1.2}
\begin{center}
\caption{Similar to Table \ref{tab:comparison_cartpole_detail}, we compare all algorithms in the Pareto Front with SAC (warm-start and hyperparameter-tuned ACME SAC) using metrics obtained in the \textbf{Gym Pendulum} environment: average performance and \generalizabilitynospace, stability-adjusted performance and \generalizability scores, and measure of instability (standard deviation $\sigma$ divided by the warm-start's $\sigma_{WS}$. We compute these metrics across 10 seeds and the best result in a \emph{column} is \textbf{bolded}. $^{\dagger}$In contrast to performance and \generalizabilitynospace, the lower the instability the better.
}
\label{tab:comparison_pendulum}
\begin{tabular}{lcc|cc|cc}
                                           & \multicolumn{2}{c|}{\textbf{Performance}} & \multicolumn{2}{c|}{\textbf{\Generalizabilitynospace} } & \multicolumn{2}{c}{\textbf{Instability}$^{\dagger}$ ($\sigma$/$\sigma_{WS}$)} \\ \cline{2-7} 
\textbf{RL Algorithm}                      & $f_{perf}$       & $\tilde{f}_{perf}$       & $f_{gen}$           & $\tilde{f}_{gen}$          & Perf.              & Gen.             \\ \hline
\textbf{Pareto 1: Best performer} & \textbf{0.887} & \textbf{0.877} & 0.360 & 0.349 & 0.45 & 0.73 \\
\textbf{Pareto 2} & 0.885 & 0.876 & 0.381 & 0.364 & \textbf{0.41} & 1.13 \\
\textbf{Pareto 3} & \textbf{0.887} & \textbf{0.877} & 0.391 & 0.377 & 0.45 & 0.93 \\
\textbf{Pareto 4} & \textbf{0.887} & 0.876 & 0.392 & 0.379 & 0.45 & 0.87 \\
\textbf{Pareto 5} & \textbf{0.887} & 0.876 & 0.393 & 0.386 & \textbf{0.41} & 0.47 \\
\textbf{Pareto 6} & 0.886 & 0.876 & 0.433 & 0.415 & 0.45 & 1.20 \\
\textbf{Pareto 7} & 0.886 & 0.875 & 0.437 & 0.418 & 0.45 & 1.27 \\
\textbf{Pareto 8: Best generalizer} & 0.868 & 0.834 & \textbf{0.445} & \textbf{0.424} & 1.55 & 1.40 \\ \hline
\textbf{Warm-start SAC} & 0.879 & 0.857 & 0.383 & 0.368 & \textit{1.0} & \textit{1.0} \\
\textbf{ACME SAC HPT Perf} & 0.880 & 0.866 & 0.392 & 0.380 & 0.64 & 0.80 \\
\textbf{ACME SAC HPT Gen} & 0.879 & 0.865 & 0.400 & 0.391 & 0.64 & \textbf{0.60}              \\ \hline
\end{tabular}
\end{center}
\end{table}

\subsection{Stability analyses for RWRL Cartpole, RWRL Walker, and Gym Pendulum}
\label{sec:stability_analyses}

\begin{figure}[!h]
\centering
\begin{subfigure}{.33\textwidth}
  \centering
  \includegraphics[width=.99\linewidth]{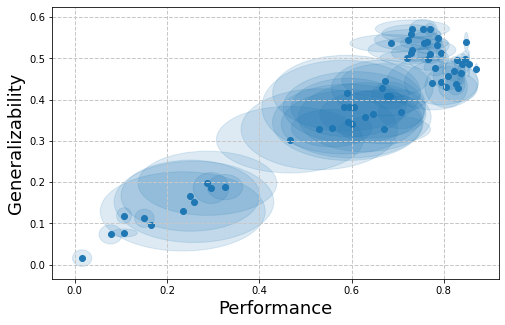}
  \caption{RWRL Cartpole}
  \label{fig:stability_cartpole}
\end{subfigure}%
\begin{subfigure}{.33\textwidth}
  \centering
  \includegraphics[width=.99\linewidth]{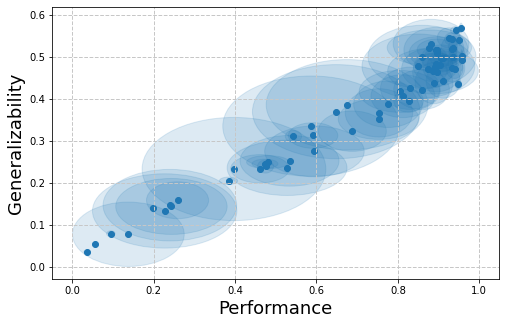}
  \caption{RWRL Walker}
  \label{fig:stability_walker}
\end{subfigure}%
\begin{subfigure}{.33\textwidth}
  \centering
  \includegraphics[width=.99\linewidth]{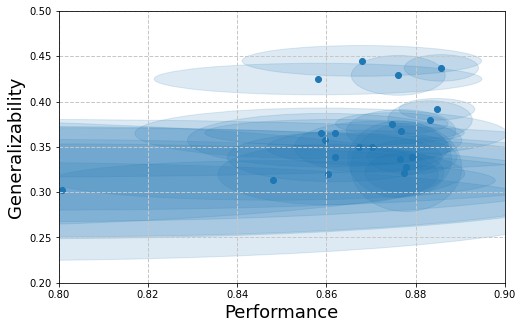}
  \caption{Gym Pendulum}
  \label{fig:stability_pendulum}
\end{subfigure}
\caption{From a subset of the meta-validated graphs, for each of them, we show the average fitness scores surrounded by an ellipse with semiaxes representing the standard deviation across seeds for each fitness score.}
\label{fig:stability}
\end{figure}

We present additional figures focused on stability, which, as introduced in Section \ref{sec:fitness}, is accounted for by penalizing the standard deviation across seeds, following Equation \ref{eq:multi-objective}. In this section, for each environment considered in this work, we select a subset of the meta-validated graphs that covers all the explored fitness space and, in Figure \ref{fig:stability}, show the average and standard deviation of each fitness score. Algorithms in the Pareto Front and those closer to it present lower variability, emphasizing that \metapg is successful in improving the stability of RL algorithms.

\subsection{Evolution results for Brax Ant}
\label{sec:additional_results_ant}

Table \ref{tab:comparison_ant} shows the meta-testing fitness scores of the best performer algorithms obtained after running evolution on Brax Ant and Brax Humanoid, respectively. We compare those against the hyperparameter-tuned warm-start SAC and observe an improvement of 15\% and 10\% in stability-adjusted performance and \generalizabilitynospace, respectively, and up to a 23\% reduction in instability. We observe that, after cross-domain transfer, the algorithm evolved in Brax Humanoid achieves comparable scores to SAC. We follow the hyperparameter tuning procedure explained in Appendix \ref{sec:hp_tuning_brax}.

\begin{table}[!h]
\def\arraystretch{1.2}
\begin{center}
\caption{Meta-tested average and stability-adjusted performance and \generalizability scores, and measure of instability (standard deviation $\sigma$ divided by the warm-start's $\sigma_{WS}$) for algorithms first evolved in Brax Ant and Brax Humanoid, and then evaluated on Brax Ant on a different set of seeds. We compare these scores against the hyperparameter-tuned warm start SAC. We compute these metrics across 4 seeds and the best result in a \emph{column} is \textbf{bolded}.}
\label{tab:comparison_ant}
\begin{tabular}{lcc|cc|cc}
                                           & \multicolumn{2}{c|}{\textbf{Performance}} & \multicolumn{2}{c|}{\textbf{\Generalizabilitynospace} } & \multicolumn{2}{c}{\textbf{Instability}$^{\dagger}$ ($\sigma$/$\sigma_{WS}$)} \\ \cline{2-7} 
\textbf{RL Algorithm}                      & $f_{perf}$       & $\tilde{f}_{perf}$       & $f_{gen}$           & $\tilde{f}_{gen}$          & Perf.             & Gen.          \\ \hline
\textbf{Ant performer} & \textbf{0.770} & \textbf{0.729} & \textbf{0.627} & \textbf{0.592} & \textbf{0.77} & \textbf{0.97} \\
\textbf{Humanoid performer} & 0.643 & 0.526 & 0.553 & 0.467 & 2.21 & 2.39 \\ \hline
\textbf{Warm-start SAC (HP tuned)} & 0.685 & 0.632 & 0.573 & 0.537 & \textit{1.0} & \textit{1.0} \\
\end{tabular}
\end{center}
\end{table}

\subsection{Evolution results for Brax Humanoid}
\label{sec:additional_results_humanoid}

\begin{figure}[!h]
\begin{center}
\includegraphics[width=0.99\textwidth]{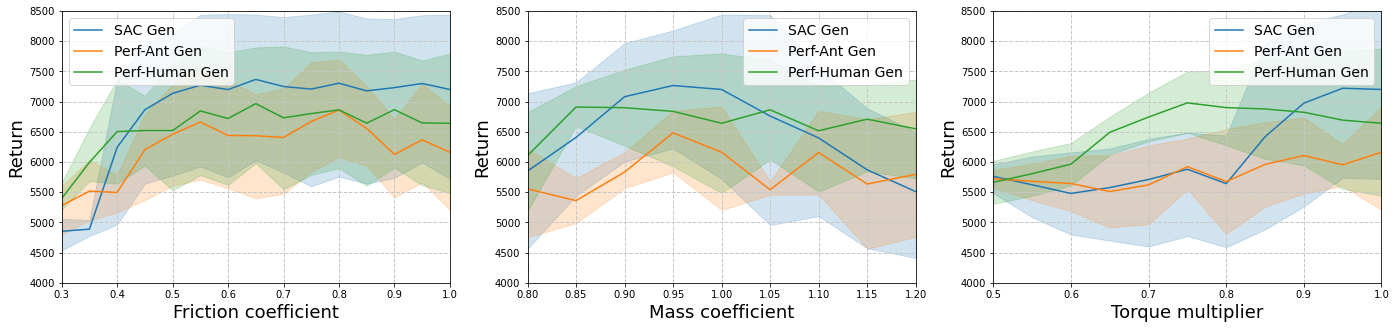}
\end{center}
\caption{Meta-testing in Brax Humanoid. We compare, after hyperparameter tuning, loss functions evolved in Brax Humanoid, Brax Ant (to assess cross-domain transfer), and the SAC baseline used as warm-start. We plot the average return and standard deviation across random seeds when evaluating in multiple Brax Humanoid instances with different friction coefficients, mass coefficients, and torque multipliers (in all cases 1.0 is used as training configuration).
}
\label{fig:perturbation_humanoid}
\end{figure}

Figure \ref{fig:perturbation_humanoid} shows the behaviour of evolved algorithms when meta-tested in Brax Humanoid, using an evaluation consisting of perturbations encountered in many practical settings (changes in friction coefficient, mass, and torque), as outlined in Appendix \ref{sec:env_configs}. We first evolve algorithms independently in both Humanoid and Ant,  then, for each case, select the algorithm with the best stability-adjusted performance $\tilde{f}_{perf}$ (in this case we focus on only one of the evolved algorithms since there is strong correlation between both metrics; the best performer and best generalizer are close, sometimes even encode the same algorithm), then meta-validate it with different hyperparameter sets (see Appendix \ref{sec:hp_tuning_brax}), and select the hyperparameter set that leads to the best stability-adjusted \generalizabilitynospace. We then fix the hyperparameters and re-evaluate using the meta-testing seeds. We compare algorithms evolved in Ant and Humanoid with hyperparameter-tuned SAC.

The numerical results of this analysis are found in Table \ref{tab:comparison_humanoid}. We observe that, while the evolved algorithm achieves better stability-adjusted \generalizability and a clear reduction in instability (up to 40\%), the evolved algorithm performs worse than SAC when meta-testing. In this case the algorithms are hyperparameter-tuned for \generalizabilitynospace, hence the higher score. Note we evolved less graphs in the specific case of Humanoid (50K compared to 200K evolved graphs for Brax Ant), as training a policy in Humanoid is more costly. We expect these results to improve if more algorithms are evolved in the population. Then, as in the case where we used Brax Ant as evaluation environment, we achieve a good cross-domain fitness compared to SAC, but the scores are lower. In this case, however, the algorithm evolved in Brax Ant shows more stability than SAC when both are evaluated in Humanoid.

\begin{table}[!h]
\def\arraystretch{1.2}
\begin{center}
\caption{Meta-tested average and stability-adjusted performance and \generalizability scores, and measure of instability (standard deviation $\sigma$ divided by the warm-start's $\sigma_{WS}$) for algorithms first evolved in Brax Ant and Brax Humanoid, and then evaluated on Brax Humanoid on a different set of seeds. We compare these scores against the hyperparameter-tuned warm start SAC. We compute these metrics across 4 seeds and the best result in a \emph{column} is \textbf{bolded}.}
\label{tab:comparison_humanoid}
\begin{tabular}{lcc|cc|cc}
                                           & \multicolumn{2}{c|}{\textbf{Performance}} & \multicolumn{2}{c|}{\textbf{\Generalizabilitynospace} } & \multicolumn{2}{c}{\textbf{Instability}$^{\dagger}$ ($\sigma$/$\sigma_{WS}$)} \\ \cline{2-7} 
\textbf{RL Algorithm}                      & $f_{perf}$       & $\tilde{f}_{perf}$       & $f_{gen}$           & $\tilde{f}_{gen}$          & Perf.             & Gen.            \\ \hline
\textbf{Ant performer} & 0.440 & 0.374 & 0.420 & 0.384 & \textbf{0.63} & \textbf{0.51} \\
\textbf{Humanoid performer} & 0.474 & 0.391 & \textbf{0.462} & \textbf{0.420} & 0.80 & 0.60 \\ \hline
\textbf{Warm-start SAC (HP tuned)} & \textbf{0.514} & \textbf{0.410} & 0.450 & 0.380 & \textit{1.0} & \textit{1.0} \\
\end{tabular}
\end{center}
\end{table}

\section{Additional analyses}
\label{sec:additional_analyses}

In this section we present additional analyses on the evolved algorithms. Specifically, we take a look at the equations defining some of the best evolved algorithms and analyze the effect of using algorithms evolved in one specific environment to train agents in different environments.

\subsection{Best performer and best generalizer for RWRL Walker and Gym Pendulum}
\label{sec:eqs_walker_pendulum}

We present the loss equations for both the best performer and best generalizer when using RWRL Walker and Gym Pendulum as training environments. First, the best performer for RWRL Walker:
\begin{align}
    L_{\pi}^{perf} &= \mathbb{E}_{(s_t, a_t, r_t, s_{t+1})\sim \mathcal{D}} \left[ r_t + \gamma\left(\underset{i}{\min}\,Q_{targ_i}(s_{t+1}, \tilde{a}_{t+1}) - \mathrm{atan}(\gamma / Q(s_t, a_t))\right) - Q(s_t, \tilde{a}_t)\right] \\
    L_{Q_i}^{perf} &= \mathbb{E}_{(s_t, a_t, r_t, s_{t+1})\sim \mathcal{D}} \left[ \left(r_t + \gamma\left(\underset{i}{\min}\,Q_{targ_i}(s_{t+1},\tilde{a}_{t+1}) - \mathrm{atan}(\gamma / Q_i(s_t, a_t))\right) - Q_i(s_t, a_t)\right)^2\right]
\end{align}
In all cases, $\tilde{a}_{t} \sim \pi(\cdot|s_{t})$, $\tilde{a}_{t+1} \sim \pi(\cdot|s_{t+1})$, and $\mathcal{D}$ is a dataset of experience tuples from the replay buffer. Next, the best generalizer for RWRL Walker:
\begin{align}
    L_{\pi}^{gen} &= \mathbb{E}_{(s_t, a_t, s_{t+1})\sim \mathcal{D}} \left[  \frac{0.2\cdot \log\pi(\tilde{a}_{t+1}|s_{t+1})}{Q_i(s_{t+1},\tilde{a}_{t+1}) - 0.1\cdot\log\pi(\tilde{a}_{t+1}|s_{t+1})} - \underset{i}{\min}\,Q_i(s_t, \tilde{a}_{t+1}) \right]\\
    L_{Q_i}^{gen} &= \mathbb{E}_{(s_t, a_t, r_t, s_{t+1})\sim \mathcal{D}} \left[ \left(r_t + \gamma\left(Q_i(s_{t+1},\tilde{a}_{t+1}) - 0.1\cdot\log\pi(\tilde{a}_{t+1}|s_{t+1})\right) - Q_i(s_t, a_t)\right)^2 \right]
\end{align}
Now we present the best performer for Gym Pendulum:
\begin{align}
    L_{\pi}^{perf} &= \mathbb{E}_{(s_t, a_t)\sim \mathcal{D}} \left[ 2\cdot\mathrm{atan}(\log\pi(\tilde{a}_t|s_t)) - \underset{i}{\min}\,Q_i(s_t, \tilde{a}_t) \right] \\
    L_{Q_i}^{perf} &= \mathbb{E}_{(s_t, a_t, r_t, s_{t+1})\sim \mathcal{D}} \left[ \left(r_t + \gamma\left(Q_{targ_i}(s_{t+1},\tilde{a}_t) - \log\pi(\tilde{a}_t|s_t)\right) - Q_i(s_t, a_t)\right)^2 \right]
\end{align}
Finally, the equations for the best generalizer when using Gym Pendulum are:
\begin{align}
    L_{\pi}^{gen} &= \mathbb{E}_{(s_t, a_t)\sim \mathcal{D}} \left[  \log(\log\pi(\tilde{a}_t|s_t)) - \underset{i}{\min}\,Q_i(s_t, \tilde{a}_t) \right]   \\
    L_{Q_i}^{gen} &= \mathbb{E}_{(s_t, a_t, r_t, s_{t+1})\sim \mathcal{D}} \left[ \left(r_t + \gamma\left(Q_{targ_i}(s_{t+1},\tilde{a}_t) - \log(\log\pi(\tilde{a}_t|s_t))\right) - Q_i(s_t, a_t)\right)^2 \right]
\end{align}

\subsection{Best performer for Brax Ant and Brax Humanoid}
\label{sec:eqs_ant_humanoid}

We present the loss equations (policy loss and critic loss) for the best performer algorithms evolved in Brax Ant and Brax Humanoid; we focus on these two algorithms in the analyses of this paper. The loss functions for the best performer for Brax Ant are:
\begin{align}
    L_{\pi}^{perf} &= \mathbb{E}_{(s_t, a_t)\sim \mathcal{D}} \left[  \log\pi(  \tilde{a}_{t+1}|s_{t+1}   ) - \underset{i}{\min}\,Q_i(s_t, a_t) \right]   \\
    L_{Q_i}^{perf} &= \mathbb{E}_{(s_t, a_t, r_t, s_{t+1})\sim \mathcal{D}} \left[ \left|\left(r + \gamma\left(\underset{i}{\min}\,Q_{targ_i}(s_t, \tilde{a}_{t+1}) - \gamma\right) - Q_i(s_t, a_t) \right)^2\cdot C_1   \right|   \right] 
\end{align}
where
\begin{equation}
    C_1 = r_t + \gamma\cdot\left(\underset{i}{\min}\,Q_{i}(s_{t+1},\tilde{a}_{t+1}) - \gamma\right)
\end{equation}
In all cases, $\tilde{a}_{t+1} \sim \pi(\cdot|s_{t+1})$ and $\mathcal{D}$ is a dataset of experience tuples from the replay buffer. Then, the equations for the best performer evolved in Brax Humanoid are:
\begin{align}
    L_{\pi}^{perf} &= \mathbb{E}_{(s_t, a_t)\sim \mathcal{D}} \left[  \log\pi(  \tilde{a}_{t+1}|s_{t+1}   ) - \underset{i}{\min}\,Q_i(s_t, a_t) \right]   \\
    L_{Q_i}^{perf} &= \mathbb{E}_{(s_t, a_t, r_t, s_{t+1})\sim \mathcal{D}} \left[ \left(C_2 - Q_i(s_t, a_t)\right)^2 \cdot C_2  \right]
\end{align}
where
\begin{equation}
    C_2 = r_t + \gamma\left( \underset{i}{\min}\,Q_{targ_i}(s_t, \tilde{a}_{t+1}) - \log\pi(\tilde{a}_{t+1}|s_{t+1}) \right)
\end{equation}

\subsection{Additional analysis on evolved algorithms for RWRL Cartpole}
\label{sec:extended_analysis_cartpole}

\begin{figure}[!h]
\centering
\begin{subfigure}{.48\textwidth}
  \centering
  \includegraphics[width=0.95\textwidth]{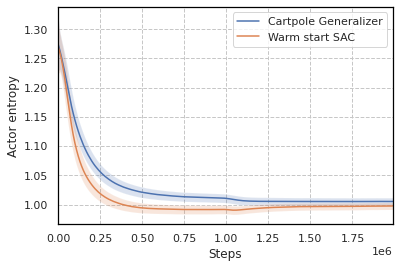}
  \caption{Average entropy of the policy during training for RWRL Cartpole.}
  \label{fig:cartpole_entropy}
\end{subfigure}%
\hspace{1em}%
\begin{subfigure}{.48\textwidth}
  \centering
  \includegraphics[width=.95\linewidth]{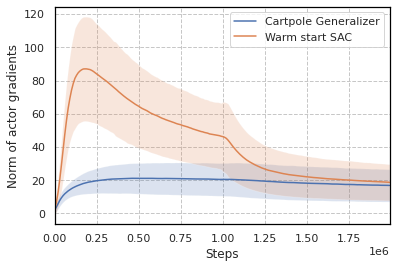}
  \caption{Average gradient norm of the actor loss during training for RWRL Cartpole.}
  \label{fig:cartpole_grad_norm}
\end{subfigure}
\caption{Analysis of the entropy and gradient norm of the actor when evaluating the best generalizer from RWRL Cartpole in comparison to the warm-start.}
\label{fig:cartpole_entropy_gradnorm}
\end{figure}

Figure \ref{fig:cartpole_entropy_gradnorm} shows the entropy and norm of the gradients of the actor for the RWRL Cartpole best generalizer. We also show these same metrics for the warm-start algorithm. This is an extension of Figure \ref{fig:cartpole_entropy_gradnorm_150000}; in both cases, we let the agents train for more episodes than those in the experimental setup. We see that ignoring this fixed number of training episodes and letting run for longer makes this type of metrics converge to similar values across algorithms. We acknowledge that training until convergence is usually preferred; however, in certain applications the number of training episodes might be a constraint, so we find \metapgnospace's ability to exploit this kind of constraints beneficial in those setups.

\subsection{Hyperparameter tuning for transfer and benchmark for RWRL and Gym}
\label{sec:transfer_procedure}

In our experiments with the RWRL Environment suite and Gym Pendulum, once an evolution experiment is over and the evolved algorithms are meta-validated, we compare them against: 1) ACME SAC \citep{hoffman2020acme}, and 2) other RL algorithms that have been evolved in a different environment. To that end, for each ACME benchmark and evolved algorithm transfer, we tune the hyperparameters of the algorithms. Since we consider two fitness scores in this work (performance and \generalizability adjusted for stability), we select the two hyperparameter configurations that lead to the best stability-adjusted performance and best stability-adjusted \generalizability scores, respectively. We denote these two configurations as the best performer and best generalizer, respectively. To that end, we do a grid search across the sets of hyperparameters listed in Table \ref{tab:hp_tuning_rwrl}.

\begin{table}[!h]
\begin{center}
\caption{Hyperparameter values considered during the tuning process.}
\label{tab:hp_tuning_rwrl}
\begin{tabular}{c|c} \hline
 
                            \textbf{Hyperparameter}       &\textbf{Values}     \\ \hline

\textbf{Discount factor} $\gamma$ &  0.9, 0.99, 0.999   \\
\textbf{Batch size} &  32, 64, 128 \\
\textbf{Learning rate} & $1\cdot10^{-4}$, $3\cdot10^{-4}$, $1\cdot10^{-3}$  \\
\textbf{Target smoothing coeff.} $\tau$ & 0.005, 0.01, 0.05 \\ 
\textbf{Reward scale} & 0.1, 1.0, 5.0, 10.0 \\ \hline

\end{tabular}
\end{center}
\end{table}

This process is only carried out once the evolution is over; the warm-start algorithm is not hyperparameter-tuned before evolution.

\subsection{Hyperparameter tuning for transfer and benchmark for Brax}
\label{sec:hp_tuning_brax}

In our experiments in Brax Ant and Brax Humanoid, given they are more costly environments, we do not meta-validate all algorithms in the population. Instead, we choose the best algorithms during meta-training and directly meta-test them with additional hyperparameter tuning. To that end, we do a grid search across the hyperparameters listed in Table \ref{tab:hp_tuning_brax} and select the configuration that maximizes the score we are interested in for each case, as described in the previous subsection.

\begin{table}[!h]
\begin{center}
\caption{Hyperparameter values considered during the tuning process.}
\label{tab:hp_tuning_brax}
\begin{tabular}{c|c} \hline
 
                            \textbf{Hyperparameter}       &\textbf{Values}     \\ \hline

\textbf{Discount factor} $\gamma$ &  0.95, 0.99, 0.999 \\
\textbf{Batch size} &  128, 256, 512 \\
\textbf{Learning rate} & $1\cdot10^{-4}$, $6\cdot10^{-4}$, $1\cdot10^{-3}$ \\
\textbf{Gradient updates per learning step}& 32, 64, 128 \\ 
\textbf{Reward scale} & 0.1, 1.0, 10.0, 100.0 \\ \hline

\end{tabular}
\end{center}
\end{table}

\subsection{Transferring algorithms evolved in RWRL and Gym}
\label{sec:transferring_algorithms}

We present the results of carrying out transfer experiments in which we take the best performer and best generalizer obtained after evolving in a specific environment (RWRL Cartpole, RWRL Walker, or Gym Pendulum) and test them in the two other environments considered. To that end, we follow the hyperparameter tuning procedure described above and therefore, for each different RL algorithm, we obtain the hyperparameter configurations that lead to the best stability-adjusted performance and best stability-adjusted \generalizabilitynospace, respectively. For example, taking the best performer from RWRL Walker (Walker Perf.) and testing it on RWRL Cartpole leads to two sets of fitness scores (best performer and best generalizer). The transfer results for RWRL Cartpole, RWRL Walker, and Gym Pendulum can be observed in Tables \ref{tab:benchmark_cartpole}, \ref{tab:benchmark_walker}, and \ref{tab:benchmark_pendulum}, respectively.

\begin{table}[!h]
\def\arraystretch{1.2}
\begin{center}
\caption{Transfer results on RWRL Cartpole. The row highlighted in gray corresponds to the results of the evolution experiment in RWRL Cartpole. The rest correspond to the best stability-adjusted performance and stability-adjusted \generalizability configurations that result from doing hyperparameter tuning to the best performer and best generalizer evolved in different environments.}
\label{tab:benchmark_cartpole}
\begin{tabular}{lcc|cc} \cline{2-5} 
\setlength{\tabcolsep}{2pt}
                        & \multicolumn{4}{c}{\textbf{RWRL Cartpole}} \\
                        & \multicolumn{2}{c}{\textbf{Best performance}}  & \multicolumn{2}{c}{\textbf{Best \generalizabilitynospace}}        \\ \hline
\textbf{RL Algorithm}   & \textbf{$\tilde{f}_{perf}$}           & \textbf{$\tilde{f}_{gen}$}  & \textbf{$\tilde{f}_{perf}$}           & \textbf{$\tilde{f}_{gen}$}      \\ \hline
\cellcolor[HTML]{EFEFEF} \textbf{Cartpole}  & \cellcolor[HTML]{EFEFEF} 0.868  &  \cellcolor[HTML]{EFEFEF} 0.459 & \cellcolor[HTML]{EFEFEF} 0.756  &  \cellcolor[HTML]{EFEFEF} 0.551 \\ \hline
\textbf{Walker Perf.} & 0.839  &  0.403 & 0.802  &  0.415 \\
\textbf{Walker Gen.} & 0.576 &  0.335 & 0.576  &  0.335 \\ \hline
\textbf{Pendulum Perf.} & 0.856  &  0.478 & 0.846  &  0.523 \\
\textbf{Pendulum Gen.} & 0.804 &  0.438 & 0.687  &  0.469 \\ \hline
\textbf{ACME SAC} & 0.864  &  0.312 & 0.833  &  0.478 \\ \hline
\end{tabular}
\end{center}
\end{table}

\begin{table}[!h]
\def\arraystretch{1.2}
\begin{center}
\caption{Transfer results on RWRL Walker. The row highlighted in gray corresponds to the results of the evolution experiment in RWRL Walker. The rest correspond to the best stability-adjusted performance and stability-adjusted \generalizability configurations that result from doing hyperparameter tuning to the best performer and best generalizer evolved in different environments.}
\label{tab:benchmark_walker}
\begin{tabular}{lcc|cc} \cline{2-5} 
\setlength{\tabcolsep}{2pt}
                        & \multicolumn{4}{c}{\textbf{RWRL Walker}} \\
                        & \multicolumn{2}{c}{\textbf{Best performance}}  & \multicolumn{2}{c}{\textbf{Best \generalizabilitynospace}}        \\ \hline
\textbf{RL Algorithm}   & \textbf{$\tilde{f}_{perf}$}           & \textbf{$\tilde{f}_{gen}$}  & \textbf{$\tilde{f}_{perf}$}           & \textbf{$\tilde{f}_{gen}$}      \\ \hline
\textbf{Cartpole Perf.} & 0.955  &  0.477 & 0.952 &  0.521 \\
\textbf{Cartpole Gen.} & 0.029  &  0.031 & 0.024  &  0.034 \\ \hline
\cellcolor[HTML]{EFEFEF} \textbf{Walker}  & \cellcolor[HTML]{EFEFEF} 0.961  &  \cellcolor[HTML]{EFEFEF} 0.526 & \cellcolor[HTML]{EFEFEF} 0.950 &  \cellcolor[HTML]{EFEFEF} 0.554  \\ \hline
\textbf{Pendulum Perf.} & 0.466  &  0.230 & 0.466  &  0.230 \\
\textbf{Pendulum Gen.} & 0.905  &  0.465 & 0.903  &  0.483 \\ \hline
\textbf{ACME SAC} & 0.965  &  0.430 & 0.918  &  0.498 \\ \hline
\end{tabular}
\end{center}
\end{table}

\begin{table}[!h]
\def\arraystretch{1.2}
\begin{center}
\caption{Transfer results on Gym Pendulum. The row highlighted in gray corresponds to the results of the evolution experiment in Gym Pendulum. The rest correspond to the best stability-adjusted performance and stability-adjusted \generalizability configurations that result from doing hyperparameter tuning to the best performer and best generalizer evolved in different environments.}
\label{tab:benchmark_pendulum}
\begin{tabular}{lcc|cc} \cline{2-5} 
\setlength{\tabcolsep}{2pt}
                        & \multicolumn{4}{c}{\textbf{Gym Pendulum}} \\
                        & \multicolumn{2}{c}{\textbf{Best performance}}  & \multicolumn{2}{c}{\textbf{Best \generalizabilitynospace}}        \\ \hline
\textbf{RL Algorithm}   & \textbf{$\tilde{f}_{perf}$}           & \textbf{$\tilde{f}_{gen}$}  & \textbf{$\tilde{f}_{perf}$}           & \textbf{$\tilde{f}_{gen}$}      \\ \hline
\textbf{Cartpole Perf.} & 0.862 &  0.329 & 0.857  &  0.349 \\
\textbf{Cartpole Gen.} & 0.821  &  0.323 & 0.821  &  0.323 \\ \hline
\textbf{Walker Perf.} & 0.860  &  0.324 & 0.813  &  0.375 \\
\textbf{Walker Gen.} & 0.849  &  0.319 & 0.679  &  0.388 \\ \hline
\cellcolor[HTML]{EFEFEF} \textbf{Pendulum}  & \cellcolor[HTML]{EFEFEF} 0.877  &  \cellcolor[HTML]{EFEFEF} 0.349 & \cellcolor[HTML]{EFEFEF} 0.834  &  \cellcolor[HTML]{EFEFEF} 0.424  \\ \hline
\textbf{ACME SAC} & 0.865  &  0.380 & 0.865  &  0.391 \\ \hline
\end{tabular}
\end{center}
\end{table}

\end{document}

%% file: math_commands.tex
\usepackage{amsmath,amsfonts,bm}

\def\eqref#1{equation~\ref{#1}}

\def\1{\bm{1}}

\DeclareMathAlphabet{\mathsfit}{\encodingdefault}{\sfdefault}{m}{sl}
\SetMathAlphabet{\mathsfit}{bold}{\encodingdefault}{\sfdefault}{bx}{n}

